# AuthorMix: Modular Authorship Style Transfer via Layer-wise Adapter Mixing

**Sarubi Thillainathan**  **Ji-Ung Lee**  **Michael Sullivan**  **Alexander Koller**
Department of Language Science and Technology, Saarland Informatics Campus
Saarland University, Saarbrücken, Germany
sarubi@lst.uni-saarland.de, ji-ung.lee@uni-saarland.de,
msullivan@lst.uni-saarland.de, koller@lst.uni-saarland.de

## Abstract

The task of authorship style transfer involves rewriting text in the style of a target author while preserving the meaning of the original text. Existing style transfer methods train a single model on large corpora to model all target styles at once: this high-cost approach offers limited flexibility for target-specific adaptation, and often sacrifices meaning preservation for style transfer. In this paper, we propose AuthorMix: a lightweight, modular, and interpretable style transfer framework. We first train individual, style-specific LoRA adapters on a small set of high-resource authors: this allows for the rapid training of specialized adaptation models for each new target using layer-wise adapter mixing via reinforcement learning, necessitating only a handful of target-style training examples. AuthorMix ranks first on the combined style–meaning score among all baselines, including GPT-5.1, and substantially improves meaning preservation over the trained baselines; under human evaluation it is the only method best-or-tied on every dimension.[1]

## 1 Introduction

Generating text that matches a targeted writing style is crucial for many applications such as personalized tutors, language learning, and conversational agents, among others (Xu et al., 2025). In these applications, appropriately adapting the style of generated text helps to foster user engagement. This is typically referred to as *authorship style transfer* and framed as a rewriting task in which a given text is rewritten to match the style of a specific target author (Patel et al., 2024; Liu et al., 2024a).

Style transfer methods face two major challenges: (i) the rewritten text must preserve the original meaning, while (ii) the style must change substantially enough to match the target style. These challenges are further compounded by the low-resource nature of authorship style transfer, as collecting a large number of texts from an individual target author is often difficult.

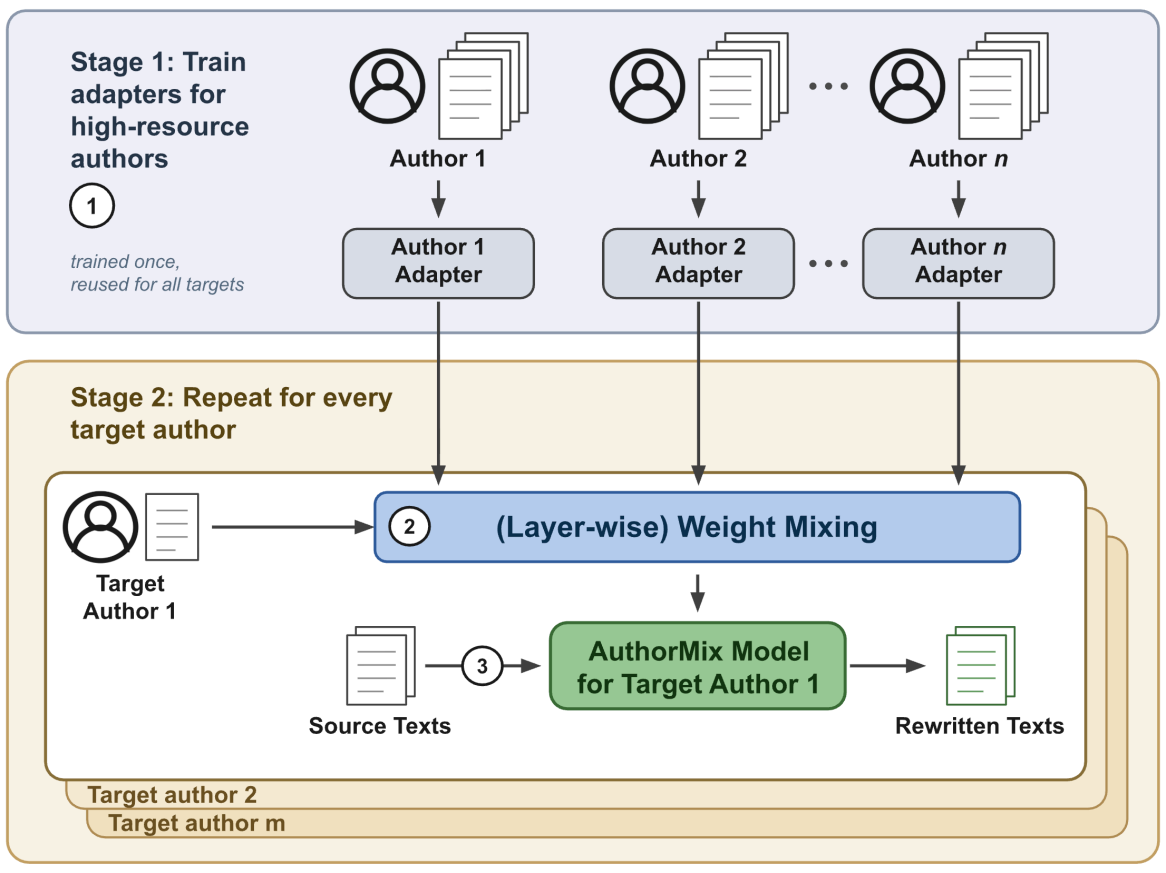

Figure 1: Overview of AuthorMix. (1) We train a LoRA adapter for each high-resource author. (2) Scalar mixing weights are optimized to jointly maximize target style and meaning preservation. (3) The resulting model rewrites any source texts into the target style.

Prior approaches address the scarcity of target-author data by using a few examples, either directly as in-context demonstrations or indirectly through learned style representations (Patel et al., 2024; Liu et al., 2024a; Horvitz et al., 2024; Riley et al., 2021). Nevertheless, transferring the target style while preserving the source meaning remains challenging (Abdulhai et al., 2026), as illustrated in Table 1.

We propose AuthorMix, a method for low-resource authorship style transfer designed to improve semantic faithfulness while maintaining target-style transfer (see Figure 1). We first train reusable LoRA adapters (Hu et al., 2022) for a small set of high-resource authors. For each new target author, we learn signed layer-wise mixing weights over these adapters using a small set of target author's examples. For this, we use reinforcement learning (RL), jointly rewarding style transfer and source meaning preservation.

[1] https://github.com/text-adaptation-saar/AuthorMix.git

| Approach | MIS | Output |
|---|---|---|
| **Source** | 1.000 | "I want to tell you something," I said: "I want to tell you all." |
| **ASTRAPOP** | 0.271 | I want to tell you,' he said. |
| **TinyStyler** | 0.800 | I wanted to tell you all, and tell you all, and tell you all, and tell you all, and tell you all, and tell you all. |
| **STYLL** | 0.731 | I have a burning desire to share with you everything, to lay it all out on the table." |
| **GPT 5.1**$_{16\text{-shot}}$ | 0.768 | I broke in at last. "I have something to say," I began; "I would lay it bare before every one of you." |
| **AuthorMix** | 0.905 | "Listen, I have something to say to you," I declared. "I have a message for all of you," I exclaimed. |
| **Neutral** | 0.830 | The speaker expressed a desire to convey information, stating that they wished to disclose everything. |

Table 1: Outputs of different style-transfer methods for a single sentence along with their mutual implication score (MIS) that captures how well the meaning is preserved. See analysis in Appendix C.11.1.

| |
|---|
| **Source:** 'I will,' he said; and instantly went off through a gate, Lizzy continuing her way. |
| **Output 1:** 'I will,' he muttered, and darted off through the gate, while Lizzy continued on her way. |
| **Output 2:** "Agreed," he exclaimed, and at once strode through the gate, while Lizzy continued on her path. |
| **Output 3**: He replied, "I will," and instantly went off through a gate, while Lizzy continued on her way. |
| **Output 4:** He replied, then swiftly exited through a gate, while Lizzy continued on her path. |
| **Output 5:** "Yes, I will," he said, and hurried through the gate, while Lizzy went on her way. |

Table 2: AuthorMix rewrites a single source sentence into different target author styles. Each output is produced by the same base model with different per-target mixing weights. Further examples, including outputs from all baseline methods, are given in Appendix C.12.

For each new target author (see Table 2), RL-based weight learning takes less than one hour; unlike prior works, the resulting model requires no target author examples at inference time. Its modular adapter library can be expanded through one-time training for additional high-resource authors and reused across future target authors.

Experiments show that AuthorMix outperforms current SoTA rewrite models in meaning preservation while matching the current SoTA in style transfer under low-resource settings. Our contributions are as follows:

1. The **first model merging approach to learn continuous, signed mixing weights with RL**, which enables dynamic per-target style adaptation from limited reference texts. We further show that layer-wise mixing with weights learnt via RL substantially outperforms static adapter-wise mixing; identifying layer-specific, gradient-based mixing as a promising direction for future work in model merging.
2. Extensive evaluations on 100 source–target author pairs, showing that **AuthorMix matches or outperforms competitive style-transfer baselines and GPT-5.1 on every primary metric**: it substantially improves meaning preservation over the trained baselines, matches few-shot GPT-5.1 on meaning at stronger style transfer, and achieves the best overall Joint score. A human evaluation study and an evaluation under held-out scoring functions support these findings.

## 2 Related Work

Jin et al. (2022) broadly categorize style transfer methods into (pseudo-)parallel and non-parallel data methods. Whereas early parallel data works heavily rely on large parallel corpora (Xu et al., 2012; Krishna et al., 2020), later works increasingly focus on low-resource scenarios with only a few available target style texts.

The lack of large parallel corpora is often mitigated via intermediate neutral style texts (i.e., pseudo-parallel data) which are then used as in-context learning examples (Patel et al., 2024) or even for policy optimization (Liu et al., 2024a). While neutral style texts compensate for a lack of parallel data, using them can lead to a semantic drift and result in lower meaning preservation; especially when they are also required during inference (and not only during training).

To avoid neutralization, non-parallel data methods often operate in a style embedding space, aiming to directly adjust the style (Rivera-Soto et al., 2021; Wegmann et al., 2022). One recent work that utilizes style embeddings is TinyStyler, that relies upon a multi-stage training pipeline to remove the reliance on neutral style text during inference (Horvitz et al., 2024).

| | STYLL | ASTRAPOP | TinyStyler | AuthorMix |
|---|---|---|---|---|
| **Method** | Few-shot prompting | Condition on few-shot SFT + DPO/CPO | Condition on Embedding SFT + self-distill | Adapter mixing |
| **Trainset authors** | – | All 10 high-resource authors | ∼1M | ∼4 high-resource authors |
| **Neutralization at inference** | Yes | Yes | No | No |
| **Adapt to new target** | Change prompt | Change prompt | Change embedding | Optimize weights |
| **Modular** | No | No | No | Yes |

Table 3: Comparison of authorship style transfer approaches. See Appendix C.10 for Training cost details.

**Adapter and model merging.** Composing independently trained adapters is an established paradigm (Pfeiffer et al., 2021; Holtermann et al., 2024). Methods differ in how the mixing weights are obtained and at what granularity they apply. AdapterSoup (Chronopoulou et al., 2023) averages domain adapters without learning any weights; LoRAHub (Huang et al., 2024) learns one scalar per LoRA (Hu et al., 2022) by gradient-free search; StyleRemix (Fisher et al., 2024) composes style-axis LoRAs with adapter-level weights using LoRAHub for authorship obfuscation; CAT (Prabhakar et al., 2025) learns a non-negative per-layer coefficient for two LoRAs by supervised gradient descent on labelled data; and RMM (Han et al., 2025) uses RL, but to merge *full models* over a *discrete* action space of source layers and merging operators. For each new target author, AuthorMix instead composes *author-style* adapters and learns *continuous, signed* layer-wise weights by RL against a non-differentiable style-and-meaning reward, so an adapter's contribution can be suppressed as well as combined.

Three common shortcomings of existing style transfer methods are that they (i) rely upon costly training (using thousands of author pairs or millions of instances), (ii) train a single monolithic model to serve all target styles, making it difficult to extend to new target styles, and (iii) remain few-shot at inference; i.e., target-style reference texts are required at every inference call. AuthorMix alleviates these issues by producing a specific target-style model in a lightweight adaptation step that does not require any few-shot examples for inference (see Table 3).

# 3 Methodology

We separate authorship style transfer into three steps, which induces modularity, eases extensibility, and allows us to produce a target-style specific model (see Figure 1):

1. **Author Adapter Training**: We train individual style adapters for a few *high resource* authors where abundant text is available.
2. **(Layer-wise) Weight Mixing**: For a given *target* author, we use a few text examples to conduct (layer-wise) weight mixing of individual style adapters.
3. **Text Rewriting**: We use the resulting model during inference to rewrite any *source* text provided by a source author.[2]

## 3.1 Author Adapter Training

We first train individual LoRA adapters (Hu et al., 2022) to rewrite source texts into the style[3] of a specific, high-resource target author $a$. Following Krishna et al. (2020), given a corpus $\mathcal{X}_a$ of texts from author $a$, we construct pseudo-parallel training pairs by paraphrasing each $x \in \mathcal{X}_a$ into a neutral version $\tilde{x}$—removing vocabulary choices, sentence rhythm, syntactic patterns—while preserving the underlying content and meaning, by prompting an LLM (see Appendix A.1). We then fine-tune a separate LoRA adapter to reconstruct the original text $x$ from the style-removed version $\tilde{x}$:

$$\mathcal{L}_{\text{SFT}} = -\frac{1}{|\mathcal{X}_a|} \sum_{x \in \mathcal{X}_a} \frac{1}{|x|} \sum_{i=1}^{|x|} \log p_{\theta+\theta_a}(x_i | \tilde{x}, x_{:i})$$

where $\theta$ are the frozen base model parameters and $\theta_a$ are the LoRA parameters for author $a$. Given $n$ high-resource authors $a_1, \ldots, a_n$, we train the set $\mathcal{A} = \{\theta^{(1)}, \ldots, \theta^{(n)}\}$ of corresponding author-specific adapters. We verify this step: the paraphrases move away from the source style (Away $= 0.91$) without drifting toward any target (Toward = 0.01), while preserving meaning (MIS = 0.79) and grammaticality (CoLA = 0.89); see Appendix A.1.1.

[2] Note that the target and source author might be the same (i.e., the user) in various real-world scenarios. However, this distinction is important for evaluation (see Section 4).

[3] By style, we refer to recurring characteristics of how an author writes, such as lexical choice, syntactic preferences, punctuation, formatting, and other surface-level linguistic patterns, rather than the underlying semantic content.

### 3.2 Weight Mixing

In our second step, we conduct a weighted linear mixing of the trained adapters $\theta^{(i)} \in \mathcal{A}$ into a single target-style adapter $\hat{\theta}$. While similar mixing approaches have been explored before with adapter-specific weights (Huang et al., 2024; Ilharco et al., 2023; Fisher et al., 2024), we mix at the finer granularity of individual layers.

For a base LLM with $L$ layers, and a given LoRA-targeted linear module $W^{(m)}$ in the $j^{th}$ layer ($1 \leq j \leq L$) of the model, let $A^{(i,m)}$, $B^{(i,m)}$ denote the LoRA weights for $W^{(m)}$ of the $i^{th}$ author adapter $\theta^{(i)}$. Then the low-rank update $\Delta W^{(m)}$ to $W^{(m)}$ from the target-style adapter is the linear combination of weights $M_{(-),j}$ across $n$ adapters, where $M \in \mathbb{R}^{n \times L}$ denotes the matrix of learnable mixing weights:

$$\Delta W^{(m)} = \sum_{i=1}^{n} M_{i,j} \cdot A^{(i,m)} B^{(i,m)}$$

This builds on the *Concatenation of LoRAs* adapter-merging method (Prabhakar et al., 2025), which also learns a coefficient per layer, but constrains it to $[0, 1]$ and composes exactly two adapters. We generalize it in two ways: $M_{i,j}$ may be negative, so an adapter can be suppressed at a given layer, and $n$ is arbitrary.

Despite the computational overhead compared to adapter-wise mixing, we conjecture that the greater expressiveness afforded by layer-wise mixing will help in accommodating author-specific characteristics that may be located at different layers. This is in line with findings in model interpretability research that attribute specific linguistic knowledge to different layers (Tenney et al., 2019; Fayyaz et al., 2021; Zhou and Srikumar, 2022). Given a low-resource target style (author) $t$ with a small set $\mathcal{X}_t$ of example texts written by $t$, we obtain $W$ by optimizing with respect to a style-transfer-specific objective function $S(x_s, x_{s\to t}, \mathcal{X}_t)$ (the *Joint Score* of Section 4.5).

For each $x_s$ in our set $\mathcal{X}_s$ of training examples, and each corresponding candidate re-write $x_{s\to t}$ of $x_s$, $S(x_s, x_{s\to t}, \mathcal{X}_t)$ is defined as the geometric mean of: (i) the *Toward Score* (Patel et al., 2024) $T(x_s, x_{s\to t}, \mathcal{X}_t)$ between the style embedding of $x_{s\to t}$ and the mean style embedding of $\mathcal{X}_t$, which measures the stylistic similarity between the candidate rewrite and the target author's example texts; and (ii) the Mutual Implication Score (Babakov et al., 2022) $MIS(x_s, x_{s\to t})$, which measures the semantic similarity between the original sentence $x_s$ and the rewrite $x_{s\to t}$ (see Section 4.5).

$$S(x_s, x_{s\to t}, \mathcal{X}_t) = \sqrt{T(x_s, x_{s\to t}, \mathcal{X}_t) \times MIS(x_s, x_{s\to t})}$$

Both components are necessary (Appendix C.9).

In both of our optimization approaches (Section 3.2.1 and Section 3.2.2), the base model and all LoRA parameters remain frozen, and only the mixing weights $W$ are updated. In practice, we restrict weight mixing to the subset of $\mathcal{A}$ corresponding to the top-$k$ most stylistically similar authors to the target style, since irrelevant adapters enlarge the search space and can degrade the merged model (adapter selection see Appendix B.4).

#### 3.2.1 LoRAHub

In our first approach, following the LoRAHub framework (Huang et al., 2024), we use Nevergrad's Auto-Optimizer (NGOpt, Rapin and Teytaud (2018)). This meta-optimizer automatically selects an appropriate gradient-free optimization strategy for the given problem.

Under this approach, the model $\pi_{W^{(i)}}$ generates a candidate rewrite $x_{s\to t}$ for each $x_s \in \mathcal{X}_s$ at each iteration $i$. The weights $W^{(i+1)}$ for the next iteration are then optimized via NGOpt to maximize the objective $\mathcal{L}_{\text{LH}}$:

$$\mathcal{L}_{\text{LH}} = \sum_{x_s \in \mathcal{X}_s} S(x_s, x_{s\to t}, \mathcal{X}_t) + \lambda \cdot |W^{(i)}|$$

Following Huang et al. (2024), we incorporate the L1 regularization term $\lambda \cdot |W^{(i)}|$ to mitigate overly large weights.

#### 3.2.2 GRPO

We additionally consider directly learning $W$ using the Group Relative Policy Optimization (GRPO; Shao et al. (2024)) reinforcement learning (RL) algorithm, with our joint score $S(x_s, x_{s\to t}, \mathcal{X}_t)$ serving as the reward function.

RL lets us optimize $S$ directly, on the model's own outputs: the score is computed over sampled rewrites, and so is non-differentiable with respect to $M$, and no gold rewrites exist for a low-resource target author to supervise against.

## 4 Experimental Setup

Our experiments include six baselines as well as four different variations of AuthorMix (adapter-and layer-wise mixing using LoRAHub and GRPO methods). We report averaged results across ten runs using different seeds for pseudo-random number generation (see Appendix B.3.2).

### 4.1 Datasets

We utilize three disjoint sets of authors with English literary text from Project Gutenberg (Gerlach and Font-Clos, 2018):

- **High-resource authors.** Liu et al. (2024b) provide a collection of 10 high-resource authors with substantial amounts of available text (more than 2,000 texts each). We use these to train our individual adapter models (see Section 3.1).
- **Target authors.** We randomly sample 10 unique authors from Project Gutenberg with 16 target texts each. This mimicks a low-resource scenario where only a small amount of text is available for a given target style.
- **Source authors.** Finally, we randomly sample another 10 unique authors from Project Gutenberg as source authors, whose texts serve as input to be style-transferred ($\mathcal{X}_s$ in Section 3.2). For each source author, we randomly split their texts into a training set of 50 texts for RL-based weight mixing (Section 3.2.2) and reserve 16 texts for testing. Following Patel et al. (2024) and Liu et al. (2024a), we sample 16 test examples for each source-target pair, resulting in $10 \times 10 = 100$ evaluation pairs and a total of 1,600 test instances.

We provide detailed statistics for all 30 authors in Appendix B.1.

### 4.2 Baselines

We compare AuthorMix against various baselines including prompting-based, embedding-conditioned, and policy-optimization methods. For all baselines, we use the same hyperparameters as reported in their respective original papers unless otherwise noted. Table 3 summarizes the key differences between different style transfer methods.

- **Neutral Text Baseline.** This serves as a lower-bound and simply rewrites source texts into a neutral style using LLaMA-3.3-70B-Instruct.
- **Few-shot Prompting.** We prompt GPT 5.1 (gpt-5.1-2025-11-13) to perform style transfer and provide 16 target style texts as in-context examples alongside the source text to be transferred.
- **STYLL.** Patel et al. (2024) propose an in-context learning approach using pseudo-parallel texts. We replicate their work using 16 in-context examples and respective style descriptors (see Appendix B.2 for more details).
- **TinyStyler.** Horvitz et al. (2024) propose a method that utilizes style embeddings during training and is capable of style transfer without requiring pseudo-parallel text during inference. We directly use their benchmarked released checkpoint. To ensure a fair comparison against single-instance baselines, we do not generate multiple candidates with a re-ranking.
- **ASTRAPOP.** Liu et al. (2024a) investigate supervised fine-tuning (SFT) as well as direct preference optimization (DPO) to train models for style transfer. Due to the lack of publicly available model weights, we re-train their approach with the same amount of data (20k samples). Specifically, we used 2,000 texts from each of the 10 high-resource authors and trained ASTRAPOP on the same base models as AuthorMix. We report results for both the SFT-only variant and the full, DPO-optimized model. We additionally re-train ASTRAPOP with AuthorMix's GRPO pipeline and reward (ASTRAPOP-GRPO), and also trained under an optimization budget matched to AuthorMix, denoted ASTRAPOP-GRPO (matched), to separate the effect of the optimizer from that of the parameterization (Appendix C.7).

### 4.3 AuthorMix Variants

AuthorMix conducts a target-specific weight mixing either via LoRAHub (Section 3.2.1) or GRPO (Section 3.2.2).

- **LoRAHub (adapter-wise)**. The adapter-wise weight mixing method proposed by Huang et al. (2024).
- **LoRAHub (layer-wise)**. The LoRAHub framework extended to layer-wise mixing.
- **GRPO (adapter-wise)**. We utilize GRPO to learn single mixing weights for each adapter.

- **GRPO (layer-wise)**. We utilize GRPO to learn layer-wise mixing weights for each adapter.

We evaluate every AuthorMix configuration for $k \in \{2, \ldots, 10\}$, where $k$ denotes the number of adapters selected (Section 3.2).

### 4.4 Base Model and Hyperparameters

We use LLaMA-3.1-8B-Instruct[4] and Qwen2.5-7B-Instruct[5] as the base models for AuthorMix. All high-resource author adapters are fine-tuned using the LLaMA-Factory framework (Zheng et al., 2024). The neutral paraphrases that are used for adapter training are generated using a LLaMA-3.3-70B-Instruct model with zero shot prompting (Appendix A.1). All experiments were conducted on a high-performance computing cluster with 5 × NVIDIA H100/A100/H200 GPUs.[6] We provide detailed hyperparameters in Appendix B.3.

### 4.5 Evaluation Metrics

Following Patel et al. (2024), our primary evaluation reports three metrics. **Toward** measures style transfer accuracy as movement toward the target author's style in STAR style-embedding space (Huertas-Tato et al., 2024), while **MIS** measures meaning preservation through bidirectional NLI entailment (Babakov et al., 2022). **Joint** is their geometric mean, Joint $= \sqrt{\text{Toward} \times \text{MIS}}$, computed for each source–target pair before averaging. Formal definitions are provided in Appendix B.5.

Since AuthorMix optimizes adapter selection and mixing weights directly against STAR style embeddings and MIS, we additionally report style transfer accuracy and meaning preservation under **held-out scoring functions**, i.e. functions that are never queried during adapter selection or weight learning. For style transfer accuracy, we compute Toward from CAV style embeddings (Wegmann et al., 2022) trained from scratch on Gutenberg text (Appendix B.5.3). For meaning preservation, we compute the cosine similarity between SBERT embeddings (Reimers and Gurevych, 2019). Agreement between the two sets of scoring functions indicates that a method's scores reflect style transfer and meaning preservation rather than fit to the particular models it was optimized against.

We also report Away (Patel et al., 2024) and CoLA (Warstadt et al., 2019) as secondary evaluation metrics in Appendix B.5.2: Away is more relevant to source-style obfuscation than to target-style transfer, while CoLA mainly serves as a fluency safeguard for adapter mixing.

### 4.6 Human Evaluation

While automated metrics provide scalable signal, they may not fully capture human perception of style, meaning, and fluency: for this reason, we complement them with a human study to validate that AuthorMix's gains hold under direct human judgement.

We conduct a human evaluation using 50 source sentences with 2 sentences per author pair across 25 source-target pairs. Each sentence was rewritten by AuthorMix (GRPO, LW $k = 4$), and all style transfer baselines toward the pair's target author, yielding 250 (source, method) items. Each item was judged by 3 raters according to meaning preservation (MP), style, and fluency. We provide the full details in Appendix C.6.

## 5 Results & Analysis

Table 4 reports results for AuthorMix and all baselines, averaged across all 100 source–target author pairs. For each AuthorMix configuration, we select the best-performing $k$ based on the primary Joint score (the optimization metric), and report the corresponding held-out scores at that same $k$; the full sweep across all $k$ is provided in Figure 4 (detailed results in Table 13). Ties are broken towards the smallest $k$, which costs least to learn (Figure 4).

Overall, **AuthorMix achieves the best Joint score under both sets of scoring functions**: layer-wise GRPO ($k = 4$) reaches 0.34 with the primary scoring functions (vs. TinyStyler 0.31, ASTRAPOP-DPO 0.29) and 0.50 with the held-out scoring functions (vs. ASTRAPOP-DPO and STYLL 0.45, TinyStyler 0.40), showing that the ranking does not depend on the scoring functions AuthorMix was optimized against. Under the primary scoring functions, paired tests put AuthorMix significantly ahead of every baseline on Joint and MIS, and statistically tied on Toward with TinyStyler and ASTRAPOP (Appendix C.2).

Re-training ASTRAPOP with AuthorMix's exact GRPO pipeline and reward — changing only

[4] meta-llama/Llama-3.1-8B-Instruct
[5] Qwen/Qwen2.5-7B-Instruct
[6] The CPU architectures varied depending on the used node.

| Method | Primary scoring functions | | | Held-out scoring functions | | |
|---|---|---|---|---|---|---|
| | Toward$_{\text{STAR}}$ ↑ | MIS ↑ | Joint$_{\text{STAR}}$ ↑ | Toward$_{\text{CAV}}$ ↑ | SBERT ↑ | Joint$_{\text{CAV}}$ ↑ |
| Neutral Text† | 0.01 ±0.04 | 0.79 ±0.10 | 0.05 ±0.10 | 0.21 ±0.19 | 0.76 ±0.04 | 0.39 ±0.17 |
| Few-shot (GPT-5.1) | 0.08 ±0.08 | 0.81 ±0.09 | 0.20 ±0.16 | 0.23 ±0.14 | 0.78 ±0.04 | 0.39 ±0.17 |
| STYLL | 0.07 ±0.09 | 0.68 ±0.10 | 0.16 ±0.15 | **0.37** ±0.22 | 0.65 ±0.06 | <u>0.45</u> ±0.20 |
| TinyStyler | 0.16 ±0.10 | 0.75 ±0.08 | <u>0.31</u> ±0.14 | 0.26 ±0.18 | 0.79 ±0.03 | 0.40 ±0.21 |
| ASTRAPOP-SFT | 0.16 ±0.11 | 0.63 ±0.12 | 0.29 ±0.13 | 0.26 ±0.22 | 0.73 ±0.06 | 0.37 ±0.22 |
| ASTRAPOP-DPO | <u>0.17</u> ±0.12 | 0.63 ±0.11 | 0.29 ±0.14 | <u>0.36</u> ±0.25 | 0.71 ±0.05 | <u>0.45</u> ±0.22 |
| ASTRAPOP-GRPO | <u>0.17</u> ±0.11 | 0.62 ±0.11 | 0.30 ±0.12 | 0.28 ±0.22 | 0.72 ±0.05 | 0.39 ±0.21 |
| ASTRAPOP-GRPO (matched) | **0.18** ±0.11 | 0.55 ±0.13 | 0.29 ±0.11 | 0.26 ±0.20 | 0.68 ±0.06 | 0.37 ±0.20 |
| *AuthorMix* | | | | | | |
| LoRAHub, AW ($k$=3) | 0.11 ±0.10 | <u>0.85</u> ±0.09 | 0.25 ±0.17 | 0.28 ±0.20 | <u>0.87</u> ±0.05 | 0.44 ±0.21 |
| LoRAHub, LW ($k$=3) | 0.12 ±0.10 | **0.86** ±0.08 | 0.27 ±0.17 | 0.27 ±0.21 | **0.89** ±0.05 | 0.42 ±0.22 |
| GRPO, AW ($k$=3) | 0.13 ±0.09 | 0.78 ±0.09 | 0.29 ±0.13 | 0.31 ±0.20 | 0.81 ±0.05 | <u>0.45</u> ±0.19 |
| GRPO, LW ($k$=4) | 0.16 ±0.10 | 0.83 ±0.07 | **0.34** ±0.12 | 0.35 ±0.20 | 0.84 ±0.05 | **0.50** ±0.20 |

Table 4: Main results over 100 source–target pairs × 10 evaluation runs; ± is the standard deviation over these 1,000 generations. ↑ = higher is better. **Bold** = best, <u>underlined</u> = second best. AW = adapter-wise, LW = layerwise mixing. †Reference bound (neutralized text, no style transfer). Both blocks report style transfer accuracy, meaning preservation, and their geometric mean; the right block uses scoring functions held out from AuthorMix's optimization (CAV in place of STAR, SBERT in place of MIS). Secondary metrics (CoLA, Away) and the full $k$-sweep are reported in Table 13.

*what* is optimized (a single monolithic model's LoRA weights, rather than mixing weights over a frozen adapter library) — does not close the gap: Joint = 0.29 at a matched 300 steps and 0.30 at 3,000 steps, versus 0.34, with far lower MIS (0.55 / 0.62 vs. 0.83; Appendix C.7).

With an identical variant of AuthorMix that restricts the mixing weights to be non-negative, as in CAT (Prabhakar et al., 2025), the Joint score falls from 0.34 to 0.01, indicating that the capacity to suppress adapters at individual layers (Figure 3) is essential (Appendix C.8). With respect to level of granularity for the weight mixing, we find that a fine-grained, layer-wise mixing consistently outperforms an adapter-wise weight mixing.

With Qwen2.5-7B-Instruct as the backbone, AuthorMix again leads under the held-out scoring functions and beats both ASTRAPOP variants, but only matches TinyStyler on the primary Joint score (Appendix C.3).

## 5.1 Additional Metrics

Both AuthorMix and TinyStyler operate directly on source text without a neutralization step, yet AuthorMix achieves a substantially higher MIS (0.83 vs. 0.75) while matching TinyStyler's Toward score (both 0.16). Methods that *do* utilize neutral text during inference (ASTRAPOP, STYLL) achieve comparable or higher Toward scores, but at the cost of lower MIS (∼0.63). While the MIS differences may appear modest numerically, Table 1 illustrates how scores around 0.75 already introduce noticeable meaning changes. Figure 2 shows the distribution of MIS scores across all instances and random seeds. AuthorMix has a consistently high MIS score with low variance (MIS $= 0.83 \pm 0.07$ the tightest spread of any method, vs. $0.63 \pm 0.11$ for ASTRAPOP-DPO and $0.75 \pm 0.08$ for TinyStyler). In contrast, all author style transfer methods either have a lower mean and median MIS and/or a larger variation. We provide detailed distributions and method-wise variances in Appendix C.4.

With respect to fluency, AuthorMix variants achieve consistently high CoLA scores of 0.83–0.87, comparable to all other baselines except both TinyStyler variants (which scores 0.68). The low fluency of TinyStyler might be attributed to its informal Reddit text training data (Horvitz et al., 2024) (in contrast to other methods trained on Project Gutenberg data) and the resulting domain mismatch might further contribute to the fluency gap when applying this model to literary text. Our failures are systematic rather than random — style under-shift on already-plain sources, clause drop on long inputs, and near-verbatim copying on verse — and are diagnosed per axis in Appendix C.12.1.

## 5.2 LoRAHub vs. GRPO

Comparing GRPO against LoraHub under identical conditions (same $k$) and similar compute budget reveals that across all configurations, **GRPO achieves higher Toward scores**; with larger dif-ferences for layer-wise mixing (vs. author-wise mixing). Although **LoRAHub variants achieve a higher MIS** (0.85–0.87 vs. 0.78–0.83), they trade-off suboptimally with the Toward score (as reflected in their lower joint scores). We can further see that GRPO benefits substantially more compared to LoRAHub from a low number $k$ of high-resource author adapters (Figure 4).

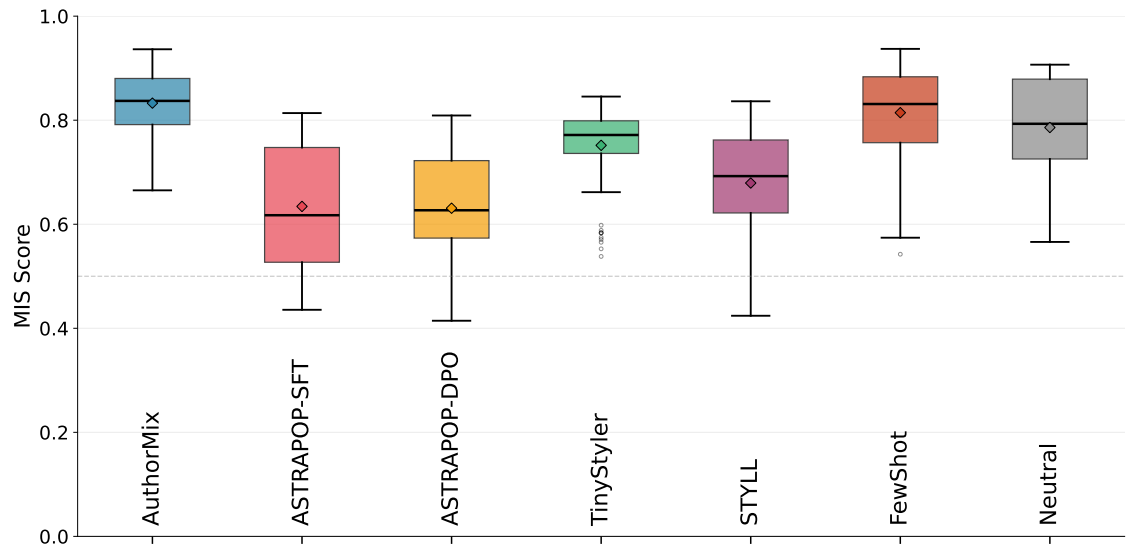

Figure 2: Distribution of MIS scores across all 100 source–target pairs for each method. Boxes show interquartile range (IQR); diamonds mark the mean; whiskers extend to 1.5×IQR. The dashed line at 0.5 indicates the threshold below which semantic equivalence breaks down.

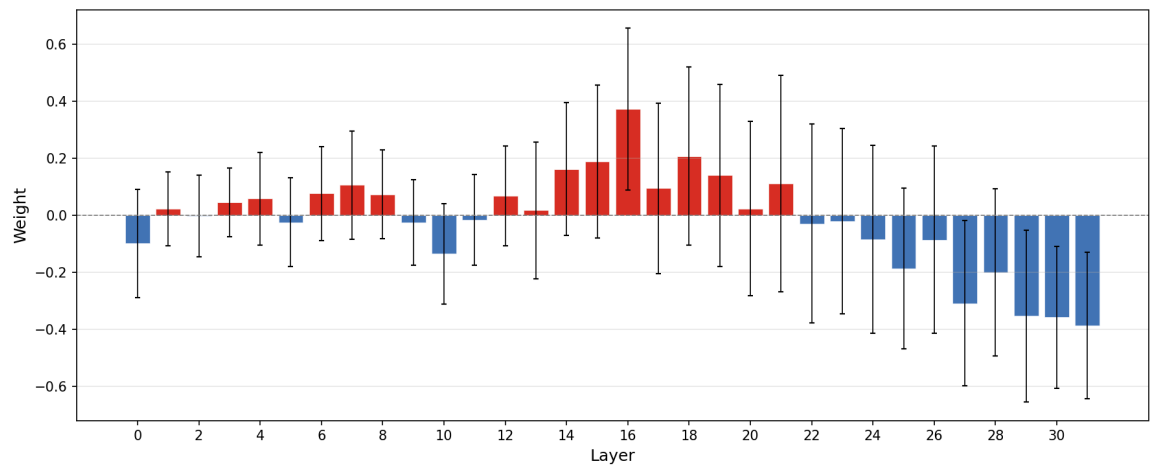

Figure 3: Global average of learned layer weights across all 10 target authors and all $k$ values for GRPO layerwise mixing. Red bars indicate positive weights (adapter contribution is used); blue bars indicate negative weights (adapter is suppressed). Vertical lines indicate $\pm 1$ standard deviation across target authors. A per-target-author heatmap is provided in Figure 12.

Finally, we analyze the individually learned weights in each layer, and visualize the averaged weights for GRPO and LoRAHub separately. Figure 3 shows the global average of learned layer weights across all 10 target authors and all $k$ configurations ($k$=2–10) for GRPO layerwise mixing. We find that GRPO learns structured layerwise patterns, with the middle layers (14–21) receiving the strongest positive weights, while early layers (0–10) show small mixed-sign weights and deep layers (24–31) are predominantly negative, suppressing adapter contributions at these levels. The high standard deviation across target authors reflects that different target authors require different layer compositions, reinforcing the need for per-target optimization. A per-target-author breakdown is provided in Figure 12.

In contrast, LoRAHub assigns near-zero weights across most layers (see Appendix C.5.1), indicating that it is not suited well for a layer-wise mixing. We conjecture that the substantially larger search space (compared to adapter-wise mixing) results in an optimization problem that is too difficult to solve well using the gradient-free methods available in NGOpt. We leave investigating a mixed-integer linear programing approach using more sophisticated solvers such as Gurobi (Gurobi Optimization, LLC, 2026) for future work.

### 5.3 Impact of Number of Adapters $k$

We conduct an ablation study to better understand the impact of using different $k$ high-resource author adapters. Figure 4 shows the results for all four AuthorMix variants.

Scores sharply rise from $k$=1 (single adapter, $\approx$ 0.17 joint score) to $k$=2 (0.23–0.32), confirming that mixing even two adapters is substantially better than using a single one. We further observe that GRPO layer-wise peaks at $k$=4 (Joint = 0.34) and remains competitive through $k$=6 before degrading at $k\geq$7. In contrast, LoRAHub variants plateau earlier around $k$=3 and remain consistent for increasing $k$. This aligns with findings by Fisher et al. (2024) who report that mixing more than five adapters hurts grammaticality and findings by Huang et al. (2024) who observe that mixing more adapters expands the search space, leading to a higher variance.

These results indicate that identifying a good subset of top-$k$ stylistically similar authors is essential, as it reduces optimization the search space and ensures that each adapter contributes a meaningful stylistic signal.

### 5.4 Human Evaluation

Table 5 shows the results of our human evaluation study. AuthorMix significantly outperforms ASTRAPOP-DPO, TinyStyler, and STYLL on **meaning preservation (MP)** (Table 16, paired $t$-tests on item means, Holm-corrected within dimension; Cohen's $d_z = 1.11, 0.75, 0.56$ respectively; all $p_{\text{Holm}} \leq 0.0005$) and ties Few-shot ($d_z = 0.18$, n.s.) matching the auto-MIS prediction (as MIS has been shown to correlate with human judgments on style transfer tasks (Babakov et al., 2022)).

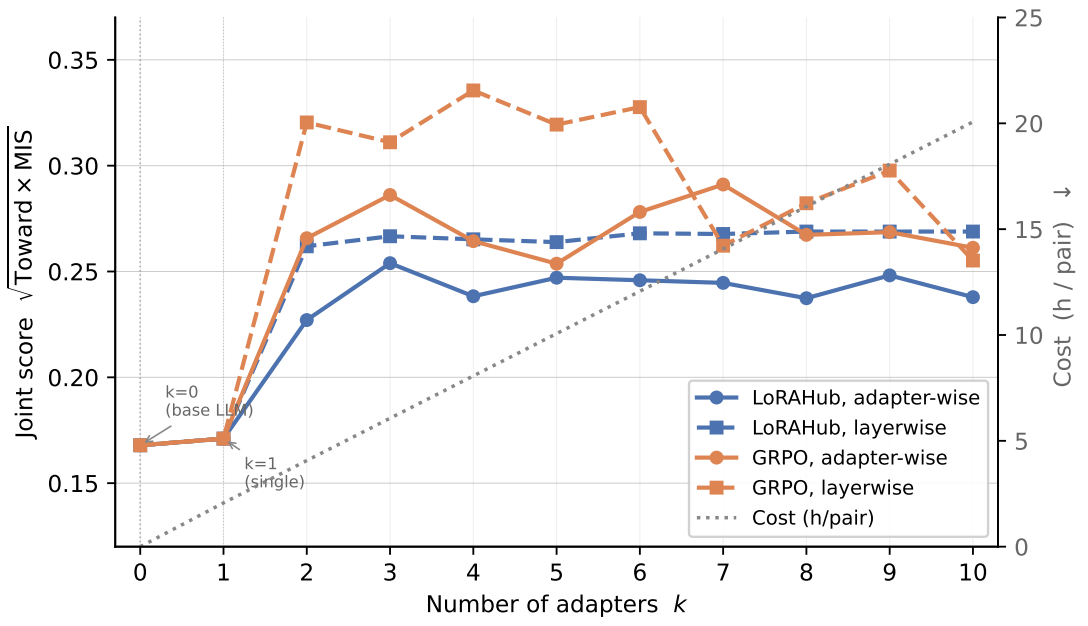

Figure 4: Joint score (left axis) and wall-clock training cost in hours per target (right axis) vs. number of related authors $k$ for all four AuthorMix variants. $k = 0$ is the base LLaMA-3.1-8B-Instruct (zero-shot, no adapter); $k = 1$ is a single best-matching adapter. All scores averaged over 100 source–target pairs.

On the **style** task, AuthorMix matches STYLL and TinyStyler but exceeds Few-shot ($d_z = 0.52$, $p_{\text{Holm}} = 0.002$); the lack of separation among the top three is consistent with the known difficulty of fine-grained authorship discrimination for untrained raters (Patel et al., 2024; Liu et al., 2024a).

On **fluency**, AuthorMix beats ASTRAPOP-DPO, Few-shot, and TinyStyler ($d_z = 0.33, 0.41, 0.94$; $p_{\text{Holm}} \leq 0.048$), and ties STYLL. Taken together, AuthorMix is the only method that is best-or-tied on every dimension — it never loses on style, meaning, or fluency — confirming under human judgement that its gains on the automatic metrics do not come at a hidden cost on any axis.

Krippendorff's $\alpha$ over the 250 items is 0.32 (meaning), 0.19 (fluency) and 0.11 (style), with raw agreement 0.51 / 0.70 / 0.56 (Table 17); comparable studies report similar levels for untrained raters (Patel et al., 2024). The ratings nonetheless validate our automatic metrics: human meaning correlates with MIS ($\rho = 0.47$) and human fluency with CoLA ($\rho = 0.41$; both $p < 0.0001$), while human style does not correlate with Toward ($\rho = 0.10$, n.s.).

### 5.5 Analyzing Resource Consumption

We further provide an analysis on resource consumption for each methods, highlighting the efficiency of AuthorMix (see Table 3). For instance, on comparable compute infrastructure, ASTRAPOP requires ∼48h of LoRA SFT + DPO training on all 10 high-resource authors (Liu et al., 2024a). TinyStyler is pre-trained using ∼8M texts from over 1M users (Horvitz et al., 2024). In contrast, AuthorMix selects a small, targeted subset of authors and adapts to any new target by learning a total of only ∼128 (k=4) scalar mixing weights which takes ∼40 minutes on a single GPU and requires no model retraining. Moreover, new adapters can be added independently to $\mathcal{A}$ with LoRA SFT requiring ∼2h of training on a single GPU.

| **Method** | **MP ↑** (0–2) | **Style acc. ↑** (vs. 0.5) | **Fluency ↑** (0–1) |
|---|---|---|---|
| Few-shot | 1.51 | 0.40 | 0.77 |
| STYLL | 1.27 | **0.63** | 0.87 |
| TinyStyler | 1.13 | 0.61 | 0.49 |
| ASTRAPOP-DPO | 0.99 | 0.57 | 0.79 |
| **AuthorMix** | **1.61** | **0.63** | **0.89** |

Table 5: Human-evaluation means score. **Bold** = best per column. Per-contrast paired-$t$ statistics ($d_z$ and Holm-corrected $p$-values) are in Table 16.

## 6 Conclusion

We introduced AuthorMix, a modular framework for low-resource authorship style transfer that composes a target style specific LoRA adapter via optimized mixing weights. Unlike prior methods that train a single, monolithic model on large corpora, AuthorMix builds a modular adapter library from a small number of high-resource authors and dynamically adapts them to any new target.

Our approach attains the best joint style–meaning score of all methods compared, and is the only one best-or-tied on every axis under both automatic and human evaluation. Our analysis shows that our novel GRPO-based method for learning a weight mixing results in interpretable, layer-wise weights; revealing that most stylistically important information is centered around the middle layers.

In future work, we plan to explore more sophisticated solvers for gradient-free weight mixing (as the ones implemented in LoRAHub failed to find meaningful weights). Moreover, the gains from layer-wise mixing suggest further untapped potential in finer-grained mixing, which also yields more interpretable weights. We believe that studying more advanced model merging strategies such as TIES-Merging (Yadav et al., 2023) or learned gating mechanisms may yield further improvements and are directly compatible with our framework. Since AuthorMix is modular, it naturally extends to cross-task style transfer by combining style adapters with task-specific ones (e.g., for summarization or dialogue) to produce stylized outputs for tasks beyond paraphrasing, and to scale larger, more diverse adapter libraries.

## Limitations

Our evaluation is conducted on a single domain (Project Gutenberg literary texts); generalization to domains with substantially different stylistic characteristics such as informal social media text or technical writing remains untested. All authors in our dataset are nonfiction/fiction writers from across different time periods; the method's effectiveness for authors with highly distinctive or contemporary styles remains to be validated.

To our knowledge, there is no widely adopted non-literary benchmark for authorship style transfer with sufficient per-author data for both adapter training and target-style references. Due to data availability and licensing constraints (e.g., Reddit), large-scale evaluation outside literary corpora is challenging. Instead we emphasized that the cross-model results already provide evidence of generalization beyond a single setup.

Authors are treated as stylistically homogeneous. Following prior work on authorship style transfer (Patel et al., 2024; Liu et al., 2024a), we treat each high-resource author as a single style label and train one adapter on all of their available texts. In practice an author's style varies across works, genres, topics and periods, so such an adapter may capture an averaged style. Our data nonetheless supports the assumption at the level we use it: under the topic-controlled CAV encoder (Appendix B.5.3), mean within-author similarity far exceeds between-author similarity (0.37 vs. 0.05). For target authors the issue does not arise, since the target style is defined by the 16 reference texts rather than an author label. Finer-grained adapters—period- or genre-specific—are a natural extension we leave to future work.

The adapter library currently contains 10 high-resource authors; scaling it would widen style coverage but increase the one-time training cost ($\sim$ 2 hrs per author), which is paid once and reused across all future targets. Each new target additionally costs less than an hour of weight learning, also paid once: the mixed adapters are then merged into the base model and no reference examples are needed at inference, whereas prompt-based baselines incur that overhead on every call. Inference cost remains unchanged across all AuthorMix variants, regardless of $k$, and is therefore comparable to baselines built on the same LLaMA-3.1-8B-Instruct backbone.

Finally, Toward is the standard metric in this literature (Patel et al., 2024), but like every embedding-based style metric (STAR, LUAR-MUD (Rivera-Soto et al., 2021), CAV (Wegmann et al., 2022)) it is not guaranteed to be fully content-independent, and disentangling style from content remains open. The Neutral Text row of Table 13 shows very low Toward alongside high MIS, indicating that the metric tracks stylistic rather than content variation.

## Ethical Considerations

Authorship style transfer carries ethical risks, as models that reproduce a person's writing style could be used for impersonation, deception, or other forms of misuse. In this work, we study the task in the context of publicly available literary texts and position it as a research problem aimed at understanding and improving personalized text adaptation, rather than enabling unauthorized imitation. Nevertheless, improved style transfer quality may also increase misuse potential. We therefore emphasize that any downstream deployment should include safeguards such as transparency about generated content, restrictions on modeling private or living individuals without consent, and careful consideration of potential harms related to identity, authorship, and misuse.

## Acknowledgments

We thank all the reviewers and our colleagues for their constructive feedback throughout this work. We are especially grateful to Isidora Jeknić for her extensive help with the output analysis and for reviewing the paper, and to Nipun Theekshana for setting up the human evaluation website. The work was funded by the Deutsche Forschungsgemeinschaft (DFG, German Research Foundation) under the project number 232722074 – SFB 1102 and in part by GRK 2853/1 "Neuroexplicit Models of Language, Vision, and Action" - project number 471607914.

## References

Marwa Abdulhai, Isadora White, Yanming Wan, Ibrahim Qureshi, Joel Z. Leibo, Max Kleiman-Weiner, and Natasha Jaques. 2026. How LLMs Distort Our Written Language. arXiv: 2603.18161 [cs.CL].

Nikolay Babakov, David Dale, Varvara Logacheva, and Alexander Panchenko. 2022. A large-scale computational study of content preservation measures for text style transfer and paraphrase generation. In *Proceedings of the 60th Annual Meeting of the Association* *for Computational Linguistics: Student Research Workshop*, pages 300–321, Dublin, Ireland.

Alexandra Chronopoulou, Matthew Peters, Alexander Fraser, and Jesse Dodge. 2023. AdapterSoup: Weight Averaging to Improve Generalization of Pretrained Language Models. In *Findings of the Association for Computational Linguistics: EACL 2023*, pages 2054–2063, Dubrovnik, Croatia.

Kevin Clark, Urvashi Khandelwal, Omer Levy, and Christopher D. Manning. 2019. What Does BERT Look at? An Analysis of BERT's Attention. In *Proceedings of the 2019 ACL Workshop BlackboxNLP: Analyzing and Interpreting Neural Networks for NLP*, pages 276–286, Florence, Italy.

Mohsen Fayyaz, Ehsan Aghazadeh, Ali Modarressi, Hosein Mohebbi, and Mohammad Taher Pilehvar. 2021. Not All Models Localize Linguistic Knowledge in the Same Place: A Layer-wise Probing on BERToids' Representations. In *Proceedings of the Fourth BlackboxNLP Workshop on Analyzing and Interpreting Neural Networks for NLP*, pages 375–388, Punta Cana, Dominican Republic.

Jillian Fisher, Skyler Hallinan, Ximing Lu, Mitchell L Gordon, Zaid Harchaoui, and Yejin Choi. 2024. StyleRemix: Interpretable Authorship Obfuscation via Distillation and Perturbation of Style Elements. In *Proceedings of the 2024 Conference on Empirical Methods in Natural Language Processing*, pages 4172–4206, Miami, Florida, USA.

Moemen Gaafar, Anurag Sarkar, and Matthew Guzdial. 2025. Few-shot Style-Conditioned LLM Text Generation via Latent Interpolation.

Martin Gerlach and Francesc Font-Clos. 2018. A standardized Project Gutenberg corpus for statistical analysis of natural language and quantitative linguistics. arXiv: 1812.08092 [cs.CL].

Gurobi Optimization, LLC. 2026. Gurobi Optimizer Reference Manual.

Jiaqi Han, Jingwen Ye, Shunyu Liu, Haofei Zhang, Jie Song, Zunlei Feng, and Mingli Song. 2025. Reinforced Model Merging. In *2025 IEEE International Conference on Multimedia and Expo (ICME)*, pages 1-6.

Carolin Holtermann, Markus Frohmann, Navid Rekabsaz, and Anne Lauscher. 2024. What the Weight?! A Unified Framework for Zero-Shot Knowledge Composition. In *Findings of the Association for Computational Linguistics: EACL 2024*, pages 1138–1157, St. Julian's, Malta.

Zachary Horvitz, Ajay Patel, Kanishk Singh, Chris Callison-Burch, Kathleen McKeown, and Zhou Yu. 2024. TinyStyler: Efficient Few-Shot Text Style Transfer with Authorship Embeddings. In *Findings of the Association for Computational Linguistics: EMNLP 2024*, pages 13376–13390, Miami, Florida, USA.

Edward J Hu, shen, Phillip Wallis, Zeyuan Allen-Zhu, Yuanzhi Li, Shean Wang, Lu Wang, and Weizhu Chen. 2022. LoRA: Low-Rank Adaptation of Large Language Models. In *International Conference on Learning Representations*.

Chengsong Huang, Qian Liu, Bill Yuchen Lin, Tianyu Pang, Chao Du, and Min Lin. 2024. LoraHub: Efficient Cross-Task Generalization via Dynamic LoRA Composition. In *First Conference on Language Modeling*.

Javier Huertas-Tato, Alejandro Martín, and David Camacho. 2024. Understanding writing style in social media with a supervised contrastively pre-trained transformer. *Knowledge-Based Systems* 296:111867.

Gabriel Ilharco, Marco Tulio Ribeiro, Mitchell Wortsman, Suchin Gururangan, Ludwig Schmidt, Hannaneh Hajishirzi, and Ali Farhadi. 2023. Editing models with task arithmetic. In *The Eleventh International Conference on Learning Representations* .

Di Jin, Zhijing Jin, Zhiting Hu, Olga Vechtomova, and Rada Mihalcea. 2022. Deep Learning for Text Style Transfer: A Survey. *Computational Linguistics* 48(1):155–205.

Kalpesh Krishna, John Wieting, and Mohit Iyyer. 2020. Reformulating Unsupervised Style Transfer as Paraphrase Generation. In *Proceedings of the 2020 Conference on Empirical Methods in Natural Language Processing (EMNLP)*, pages 737–762, Online.

Shuai Liu, Shantanu Agarwal, and Jonathan May. 2024a. Authorship Style Transfer with Policy Optimization. arXiv: 2403.08043 [cs.CL].

Xinyue Liu, Harshita Diddee, and Daphne Ippolito. 2024b. Customizing Large Language Model Generation Style using Parameter-Efficient Finetuning. In *Proceedings of the 17th International Natural Language Generation Conference*, pages 412–426, Tokyo, Japan.

Ajay Patel, Nicholas Andrews, and Chris Callison-Burch. 2024. Low-Resource Authorship Style Transfer: Can Non-Famous Authors Be Imitated? arXiv: 2212.08986 [cs.CL].

Ajay Patel, Jiacheng Zhu, Justin Qiu, Zachary Horvitz, Marianna Apidianaki, Kathleen McKeown, and Chris Callison-Burch. 2025. StyleDistance: Stronger Content-Independent Style Embeddings with Synthetic Parallel Examples. In *Proceedings of the 2025 Conference of the Nations of the Americas Chapter of the Association for Computational Linguistics: Human Language Technologies (Volume 1: Long Papers)*, pages 8662–8685, Albuquerque, New Mexico.

Jonas Pfeiffer, Aishwarya Kamath, Andreas Rücklé, Kyunghyun Cho, and Iryna Gurevych. 2021.

AdapterFusion: Non-Destructive Task Composition for Transfer Learning. In *Proceedings of the 16th Conference of the European Chapter of the Association for Computational Linguistics: Main Volume*, pages 487–503, Online.

Akshara Prabhakar, Yuanzhi Li, Karthik Narasimhan, Sham Kakade, Eran Malach, and Samy Jelassi. 2025. LoRA Soups: Merging LoRAs for Practical Skill Composition Tasks. In *Proceedings of the 31st International Conference on Computational Linguistics: Industry Track*, pages 644–655, Abu Dhabi, UAE.

J. Rapin and O. Teytaud. 2018. Nevergrad - A gradient-free optimization platform. https://GitHub.com/FacebookResearch/Nevergrad.

Nils Reimers and Iryna Gurevych. 2019. Sentence-BERT: Sentence Embeddings using Siamese BERT-Networks. In *Proceedings of the 2019 Conference on Empirical Methods in Natural Language Processing and the 9th International Joint Conference on Natural Language Processing (EMNLP-IJCNLP)*, pages 3982–3992, Hong Kong, China.

Parker Riley, Noah Constant, Mandy Guo, Girish Kumar, David Uthus, and Zarana Parekh. 2021. TextSETTR: Few-Shot Text Style Extraction and Tunable Targeted Restyling. In *Proceedings of the 59th Annual Meeting of the Association for Computational Linguistics and the 11th International Joint Conference on Natural Language Processing (Volume 1: Long Papers)*, pages 3786–3800, Online.

Rafael A. Rivera-Soto, Olivia Elizabeth Miano, Juanita Ordonez, Barry Y. Chen, Aleem Khan, Marcus Bishop, and Nicholas Andrews. 2021. Learning Universal Authorship Representations. In *Proceedings of the 2021 Conference on Empirical Methods in Natural Language Processing*, pages 913–919, Online and Punta Cana, Dominican Republic.

Zhihong Shao, Peiyi Wang, Qihao Zhu, Runxin Xu, Junxiao Song, Xiao Bi, Haowei Zhang, Mingchuan Zhang, Y. K. Li, Y. Wu, and Daya Guo. 2024. DeepSeekMath: Pushing the Limits of Mathematical Reasoning in Open Language Models. arXiv: 2402.03300 [cs.CL].

Ian Tenney, Dipanjan Das, and Ellie Pavlick. 2019. BERT Rediscovers the Classical NLP Pipeline. In *Proceedings of the 57th Annual Meeting of the Association for Computational Linguistics*, pages 4593–4601, Florence, Italy.

Alex Warstadt, Amanpreet Singh, and Samuel R. Bowman. 2019. Neural Network Acceptability Judgments. *Transactions of the Association for Computational Linguistics* 7:625–641.

Anna Wegmann, Marijn Schraagen, and Dong Nguyen. 2022. Same Author or Just Same Topic? Towards Content-Independent Style Representations. In *Proceedings of the 7th Workshop on Representation Learning for NLP*, pages 249–268, Dublin, Ireland.

Wei Xu, Alan Ritter, Bill Dolan, Ralph Grishman, and Colin Cherry. 2012. Paraphrasing for Style. In *Proceedings of COLING 2012*, pages 2899–2914, Mumbai, India.

Yiyan Xu, Jinghao Zhang, Alireza Salemi, Xinting Hu, Wenjie Wang, Fuli Feng, Hamed Zamani, Xiangnan He, and Tat-Seng Chua. 2025. Personalized Generation In Large Model Era: A Survey. In *Proceedings of the 63rd Annual Meeting of the Association for Computational Linguistics (Volume 1: Long Papers)*, pages 24607–24649, Vienna, Austria.

Prateek Yadav, Derek Tam, Leshem Choshen, Colin Raffel, and Mohit Bansal. 2023. TIES-Merging: Resolving Interference When Merging Models. In *Thirty-seventh Conference on Neural Information Processing Systems*.

Yaowei Zheng, Richong Zhang, Junhao Zhang, Yanhan Ye, and Zheyan Luo. 2024. LlamaFactory: Unified Efficient Fine-Tuning of 100+ Language Models. In *Proceedings of the 62nd Annual Meeting of the Association for Computational Linguistics (Volume 3: System Demonstrations)*, pages 400–410, Bangkok, Thailand.

Yichu Zhou and Vivek Srikumar. 2022. A Closer Look at How Fine-tuning Changes BERT. In *Proceedings of the 60th Annual Meeting of the Association for Computational Linguistics (Volume 1: Long Papers)*, pages 1046–1061, Dublin, Ireland.

| Measurement | Value | Interpretation |
|---|---|---|
| Away from source style (STAR) | 0.91 | 91% of the way from source to a random target, strong style removal |
| Toward target style (STAR) | 0.01 | near-zero residual bias toward any specific author |
| MIS (NLI bidirectional) | 0.79 | meaning preserved well above the 0.5 semantic-equivalence threshold |
| CoLA fluency | 0.89 | fluency preserved at the highest level among all evaluated outputs |

Table 6: Neutralization quality measurements aggregated over 100 source–target pairs. **Away** = 0.91 confirms that the paraphrase has been pushed far away from the source author's embedding region; **Toward** = 0.01 confirms it lands in roughly neutral territory rather than drifting toward any particular target; **MIS** = 0.79 confirms that meaning is largely retained; **CoLA** = 0.89 confirms grammaticality. The Away/Toward asymmetry (0.91 vs. 0.01) shows the paraphrase actively removes source style without injecting any particular author's style, the behavior required for pseudo-parallel adapter training.

# A Methodology

## A.1 Pseudo-Parallel Data Generation

For each high-resource author's text, we generate a neutral paraphrase using LLaMA-3.3-70B-Instruct with the prompt: *"Paraphrase the following text into a neutral, encyclopedic tone, removing stylistic markers while keeping meaning unchanged."* The resulting (neutral, original) pairs serve as supervised fine-tuning data for the high-resource author adapters.

### A.1.1 Neutralization Quality

Since parallel authorship style-transfer data is not available, we adopt the STRAP-style paraphrase-based construction of pseudo-parallel data (Krishna et al., 2020) to train the high-resource author adapters. We conduct additional experiments to evaluate how successfully the LLaMA-3.3-70B removed the style. This is important in order to ensure that the high-resource adapters are not only learning to reconstruct the content. We use the same neutralization prompt on the held-out source-author texts and score those neutral paraphrases using the same evaluation pipeline as in our main experiments (Table 13, **Neutral Text** † row). This results in four direct measurements averaged over all 100 pairs (10 source authors × 10 target authors) as shown in Table 6 and which show the successful neutralization for following reasons. First, if the neutralization were merely a synonym swap that preserved authorial style, Away would remain near 0; the observed 0.91 indicates that 91% of the embedding-space distance to a random target author has been traversed purely by stripping source style, without any per-target conditioning. Second, if the neutralization were drifting the text toward some default style (e.g., the LLaMA-3.3 chat default leaking into the output), Toward against the selected target author would deviate from 0; the observed 0.01 is essentially indistinguishable from zero, ruling this out. Finally, the high MIS = 0.79 and CoLA = 0.89 rule out the alternative failure mode where neutralization removes style by destroying the sentence; the paraphrase is grammatical and meaning-preserving.

| Author group (#) | Train | Test |
|---|---|---|
| High-resource (10) | > 2,000 | — |
| Source (10) | 50 | 16 |
| Target (10, low-resource) | — | 16 refs |

Table 7: Per-author sentences. High-resource authors are used only for adapter SFT (neutral→original pairs). Source-author Train sentences drive weight optimization; Test sentences are paired with all 10 targets at evaluation (100 pairs, 1,600 instances). Target authors are low-resource: only the 16 reference sentences are ever used (as the Toward-reward target).

The same prompt is used during both high-resource adapter training (neutral → original recovery) and the Neutral-Text test-time baseline, so this measurement directly verifies the behavior of the procedure that produces the SFT data for the adapter library.

# B Experimental Setup

## B.1 Author Lists

Project Gutenberg is an open repository of literary works by authors whose writings have entered the public domain[7]. Its eBooks may be used for a wide range of purposes, including the creation of derivative works, reports, performances, and research. All texts used in this work were obtained from Project Gutenberg and are drawn from English-language authors.

**High-resource (adapter) authors.** Mark Twain, Virginia Woolf, Vernon Lee, Charlotte Perkins Gilman, George Orwell, Jane Austen,

[7] https://www.gutenberg.org/policy/license.html

Nathaniel Hawthorne, Oscar Wilde, P. G. Wodehouse, and Samuel Richardson (obtained from (Liu et al., 2024b)[8]).

**Low-resource target authors.** Arthur Conan Doyle, Joseph Conrad, Jack London, E. M. Forster, F. Scott Fitzgerald, Rudyard Kipling, Robert W. Chambers, Agatha Christie, Edith Wharton, and L. M. Montgomery.

**Source authors.** L. Frank Baum, Charlotte Brontë, Lewis Carroll, Charles Dickens, Thomas Hardy, G. A. Henty, John Ruskin, Walter Scott, Henry B. Wheatley, and David Widger.

#### B.1.1 Data Splits per Author

Table 7 lists the per-author texts used at each stage. All splits are disjoint, and the three author groups are mutually disjoint.

### B.2 Baseline Details

We evaluate STYLL on GPT-3.5turbo since the GPT-3 endpoint used in (Patel et al., 2024) is deprecated by OpenAI, and GPT-3.5-turbo is the closest available model. We did not include BLOOM-7B based on STYLL, since MIS is in low range.

### B.3 Base Model and Hyperparameters

#### B.3.1 High-Resource Adapter Training

All adapters are fine-tuned using LLaMA-Factory (Zheng et al., 2024) with the following configuration: LoRA rank 8, $\alpha = 16$, applied to all linear target modules, input cutoff length of 256 tokens, and maximum output length of 256 tokens. *high-resource author* adapters and AuthorMix trained using following template: Listing 1 (no other additional input like previous works). For weight optimization, merging, and inference we largely follow the hyperparameter choices of StyleRemix (Fisher et al., 2024) (e.g., top-$p = 0.95$, temperature $\tau = 1.0$). All training (AuthorMix and baselines) is done using a fixed seed of 42. For all evaluation (across the paper), we report results averaged across ten runs with different random seeds.

#### B.3.2 Stability Across Training Seeds

We distinguish two sources of randomness. **Evaluation randomness** comes from sampling-based decoding; every number reported in this paper is averaged over ten evaluation runs with different decoding seeds, and the corresponding distributions and standard deviations are given in Appendix C.4. **Training randomness** enters AuthorMix only through GRPO rollout sampling, as the mixing weight matrix $M$ is initialized deterministically at zero (Section 3.2.2). Re-running GRPO layer-wise ($k = 4$) with three training seeds, over all 10 target authors and 100 source–target pairs, gives Toward $= 0.16 \pm 0.004$, MIS $= 0.83 \pm 0.010$ and Joint $= 0.33 \pm 0.003$ (mean $\pm$ standard deviation across seeds) — an order of magnitude below the gaps to the baselines in Table 4.

**Per-author SFT hyperparameters.** One LoRA adapter is fine-tuned independently for each of the 10 high-resource authors on its own neutral $\rightarrow$ original passages, yielding a library of 10 author-specific adapters. The hyperparameters in Table 9 are identical across all 10 authors.

```
{
    "instruction": "Paraphrase",
    "input": "By 2020 and beyond this
number may have increased even further.",
    "output": "it is possible that this
number will increase even later by 2020."
}
```

Listing 1: Example input-output pair used in adapter fine-tuning for AuthorMix.

#### B.3.3 Prompt Templates Across Methods

One possible reason why the performances vary across methods may be the different prompts that have been used. Table 8 shows all prompts used across different methods. Most notably, **AuthorMix uses the simplest prompt**: a one-word instruction "Paraphrase" with the source text as the only input. Moreover, multiple baselines (Few-shot, STYLL, ASTRAPOP) explicitly embed target information in the prompt, consuming substantially more context (5–16 reference texts of $\sim$ 25–150 tokens each) which substantially increases the per-call cost. This further highlights the superiority of AuthorMix.

### B.4 Adapter Subset Selection

While weight mixing can be conducted using all adapters available in $\mathcal{A}$, this does not scale well with an increasing number of adapters, especially for gradient-free mixing, which often relies on search-based algorithms. Moreover, considering irrelevant adapters for weight mixing can even hurt the performance of the resulting model (see our ablation study in Section 5.3). To address this, we restrict to the subset of $\mathcal{A}$ corresponding to the top-$k$ most stylistically similar authors to the target style for weight mixing. We quantify style similarity using the cosine similarity between prototypical style embeddings $e_a$ for high-resource author $a$ and target style embeddings $e_t$. We compute $e_a$ by randomly sampling a set of texts $\mathcal{X}'_a \subset \mathcal{X}_a$ and averaging the resulting embeddings:

$$e_a = \frac{1}{|\mathcal{X}'_a|} \sum_{i=1}^{|\mathcal{X}'_a|} \text{Emb}(x_i)$$

[8] https://github.com/cauchy221/StyleTunedLM

| Method | #in-ctx examples | Input is | Prompt structure |
|---|---|---|---|
| AuthorMix | 0 | source | {"instruction": "Paraphrase", "input": <source>} — target style is delivered by the merged LoRA adapter, not by the prompt. |
| Few-shot (GPT-5.1) | 16 | source | "Paraphrase the following text in the style of the examples below." followed by 16 numbered target-author style texts (the full 16-text reference set), then "Now paraphrase this text: {src}\n\nParaphrase:". |
| STYLL | 16 | neutral | Replicates the original STYLL prompt (Patel et al., 2024). For each target author, the LLM first generates a variable-length list of style descriptors (e.g. "mysterious, descriptive, historical") summarising the target's writing; these are then woven into 16 paired (neutral, original) in-context exemplars formatted as "Here is some text: <neutral>\nHere is a rewrite of the text that is more <descriptors>: <original>", ending with the test query "Here is some text: {src_neutral}\nHere is a rewrite of the text that is more <descriptors>:". |
| ASTRAPOP | 5 | neutral | Reference texts wrapped in [REF]...[/REF] tags in the instruction field (5 target-author exemplars), followed by the neutral source as input; the model is SFT/DPO-tuned to emit the original, follows Liu et al. (2024a). The full prompt-with-references is part of the training data, not just inference only. |
| TinyStyler | 16* | source | The source text along with style vector (the 16 target-author reference texts are each encoded with the Style-Embedding model and **mean-pooled** into a single style vector) pass as input to the T5 model. TinyStyler generates the rewrite conditioned jointly on the encoded source and the style vector (Horvitz et al., 2024). |

Table 8: Prompt / conditioning summary across all evaluated methods. #in-ctx examples = number of target-author reference texts inserted into the prompt at inference time. **Input is** indicates whether the model receives the raw source text or a neutralised paraphrase of it. * For TinyStyler, the 16 references are encoded into a single embedding rather than serialised into the prompt.

We similarly obtain a representative embedding for the target style $t$:

$$e_t = \frac{1}{|\mathcal{X}_t|} \sum_{i=1}^{|\mathcal{X}_t|} \text{Emb}(x_i)$$

Finally, we select the top-$k$ most similar adapters based on the resulting ranking using the cosine similarity $\cos(e_t, e_a)$ between all available authors and the target style.

### B.4.1 Weight Learning Details

For each target author, we learn the layerwise mixing weights using GRPO (Section 3.2.2) on the held-out source-author texts: each candidate weight configuration generates a batch of style-transferred texts, which are scored against the target author's reference texts using the joint reward (Section 3.2.1). This is conceptually similar to the policy optimization stage of (Liu et al., 2024a) however, in our case the base model and all adapter parameters are frozen — only the $m \times L$ scalar mixing weights are tuned (e.g., $4 \times 32 = 128$ weights with $k = 4$ related authors and LLaMA-3.1-8B's 32 transformer layers). For each input, the model samples a group of $G$ candidate outputs, and the mixing weights are updated using relative reward rankings within each group.

### B.4.2 Gradient-Free Optimization Hyperparameters

For the gradient-free learning (Section 4.3), we use NGOpt (Rapin and Teytaud, 2018) as provided by LoraHub (Huang et al., 2024) with the configurations provided in Table 10. We use the comparable compute budgets to match with GRPO based weight learning.

| Hyperparameter | Value |
| --- | --- |
| Base model | LLaMA-3.1-8B-Instruct |
| Stage | SFT |
| Chat template | llama3 |
| Finetuning | LoRA |
| LoRA rank | 8 |
| LoRA $\alpha$ | 16 |
| LoRA target | all linear modules |
| LoRA dropout | 0.05 |
| Cutoff length | 256 tokens |
| Optimizer | AdamW (default) |
| Learning rate | $1 \times 10^{-4}$ |
| LR scheduler | cosine |
| Warmup steps | 200 |
| Max steps | 5,000 |
| Per-device batch size | 8 |
| Gradient accumulation | 4 (effective batch = 32) |
| Precision | bf16 |

Table 9: Supervised-fine-tuning hyperparameters for the 10 high-resource author adapters. Each adapter is trained independently on its author's neutral→original pairs and stored in the adapter library $\mathcal{A}$ for later mixing (Section 3.2).

| Hyperparameter | Value |
| --- | --- |
| Optimizer | NGOpt |
| Optimization steps | 250 |
| Weight initialization | 0 |
| Weight bounds | [−1.5 to +1.5] |
| Top-$p$ sampling | 0.95 |
| Temperature $\tau$ | 1.0 |

Table 10: LoRAHub, Gradient-free optimization hyperparameters.

#### B.4.3 GRPO Hyperparameters

We use hyper parameters as given in Table 11. we set $\beta$ controls the optional KL regularization to zero.

| Hyperparameter | Value |
| --- | --- |
| Learning rate | 0.02 |
| Optimization steps | 300 |
| Top-$p$ sampling | 0.95 |
| Temperature $\tau$ | 1.0 |
| Weight initialization | 0 |

Table 11: GRPO optimization hyperparameters.

### B.5 Evaluation Metrics

#### B.5.1 Primary Metrics: Formal Definitions

Following Patel et al. (2024), we report three primary metrics:

- **Toward.** Measures how far the transferred output has moved toward the target author's style, as a fraction of the maximum possible movement. Given source text $x_s$, target author $t$, and transferred text $x_{s\to t}$ with respective style embeddings $e_s$, $e_t$, and $e_{s\to t}$:

$$\text{Toward} = \frac{\max(Sim(e_{s\to t}, e_t) - Sim(e_s, e_t), 0)}{1 - Sim(e_s, e_t)}$$

  where $Sim$ is the angular similarity $Sim(u, v) = 1 - \arccos(u \cdot v/(\|u\| \cdot \|v\|))/\pi$ over style embeddings from STAR (Huertas-Tato et al., 2024) — a RoBERTa-large encoder pre-trained on 4.5M texts from 70K authors (including authors from Project Gutenberg).[9]
- **Semantic preservation (MIS).** The mutual implication score (MIS, Babakov et al. (2022)) evaluates bidirectional entailment between the source and transferred texts using an NLI model. MIS ranges between 0 and 1 and has been shown to correlate well with human judgments of meaning preservation in style transfer tasks (Babakov et al., 2022; Patel et al., 2024).
- **Joint.** The geometric mean of Toward and MIS: $\text{Joint} = \sqrt{\text{Toward} \times \text{MIS}}$, capturing the trade-off between style transfer and meaning preservation. Following Patel et al. (2024), Joint is computed *per source–target pair* and then averaged over all 100 pairs and ten evaluation runs; it is therefore not recoverable as the geometric mean of the averaged

[9] LUAR-MUD (Rivera-Soto et al., 2021) was not suitable for our selected author set; see Gaafar et al. (2025) for discussion.

Toward and MIS columns. This per-pair protocol properly balance both axes on *every* pair, rather than letting a strong style score on one pair compensate for a collapsed meaning score on another. The same convention is used for Joint$_{\text{CAV}}$ under the held-out scoring functions.

As we use the toward, MIS, and joint scores in the optimization objective of AuthorMix but also at the same time as our primary evaluation metrics, this might favor AuthorMix in the evaluation. We thus report style transfer accuracy and meaning preservation a second time under **held-out scoring functions** (functions that AuthorMix is not optimized against) and find that AuthorMix still performs best under them.

#### B.5.2 On Away and CoLA as Secondary Metrics

**Away score.** The Away metric measures how far the transferred text has moved *away* from the source author's style, expressed as a fraction of the source–target distance (Patel et al., 2024). We exclude it from the Joint score for two reasons. First, Away is under-determined: many stylistically different outputs can achieve the same Away value, since any text that is equidistant from the source in the embedding space scores identically — regardless of whether it moved toward the target or in an unrelated direction. This makes Away a weak signal for evaluating style *transfer* quality. Second, a high Away score primarily matters in **authorship obfuscation** settings — such as those targeted by StyleRemix (Fisher et al., 2024) — where the explicit goal is to hide the source author's identity. In standard style transfer, the objective is to approach the target style (captured by Toward) while preserving content (captured by MIS); a text can succeed on both without fully erasing source-style traces, which is acceptable in most practical applications.

**CoLA score.** Large language models, especially instruction-tuned variants like LLaMA-3.1-8B-Instruct, generally produce fluent and grammatical text, making fluency a less discriminative metric in recent work. However, (Fisher et al., 2024) (who trained on Base LLaMa model) observe that adapter merging — which is central to our approach — can introduce fluency degradation, particularly when merging five or more adapters. Since AuthorMix composes multiple LoRA adapters via learned weights, we include CoLA as a safeguard to verify that the merging process does not degrade linguistic acceptability. Our results confirm that AuthorMix maintains high CoLA scores (0.83–0.86), comparable to non-merged baselines, indicating that the learned weight composition does not harm fluency.

#### B.5.3 Training of the Held-out CAV Style Encoder

To evaluate generations under a held-out style encoder that was **not** involved in adapter selection, weight learning, or the primary Toward evaluation (see Appendix C.1), we trained an in-domain CAV (Contrastive Authorship Verification) (Wegmann et al., 2022) style encoder from scratch on Project Gutenberg literary text. The encoder is referred to throughout the paper as **CAV**.

**Base model and architecture.** We use `roberta-base` as our base model and wrap it with a `sentence-transformers` mean-pooling head, resulting in 768-dimensional embeddings. For training, we use the standard Triplet Loss with cosine distance and margin 0.5.

**Training data.** We use the Gutenberg subset compiled by Liu et al. (2024b) 50 authors with $\sim$ 2,000 paragraph chunks per author (with a maximum sequence length of 256 tokens). The 50 authors form a superset that includes the 30 authors used in the main style-transfer evaluation (10 high-resource + 10 source + 10 target) together with 20 additional Gutenberg authors. We ensure that the CAV training texts are still disjoint from the development and test instances at the chunk level. For each author, we sample 2,000 training and 500 testing chunks.

**CAV triplets.** The core idea of CAV is that **style** and **content** signals can be disentangled by carefully constructing triplets so that the model is forced to ignore the topic. We follow this protocol:

- **Positive pair (anchor, positive).** Two chunks by the same author drawn from different pseudo-topic clusters.
- **Negative pair (anchor, negative).** Two chunks by different authors drawn from the same pseudo-topic cluster.

Pseudo-topics are obtained by applying $k$-means clustering with $k = 50$ clusters on `all-MiniLM-L6-v2` content embeddings, computed once over the full chunk pool. The contrastive training objective forces the model to place the same author closer together despite a topic shift, and place different authors farther apart despite a topic match—

| Encoder | R@8↑ | MRR↑ | AUC↑ | C@1↑ | sep.↑ | $d$↑ | R@1↑ |
|---|---|---|---|---|---|---|---|
| CAV (ours) | <u>58.3</u> | <u>0.32</u> | **0.72** | **0.64** | **7.18** | **0.82** | <u>18.1</u> |
| STAR (Huertas-Tato et al., 2024) | **63.7** | **0.38** | <u>0.71</u> | <u>0.57</u> | 1.09 | <u>0.79</u> | **24.5** |
| LUAR-MUD (Rivera-Soto et al., 2021) | 40.0 | 0.20 | 0.58 | 0.50 | 1.06 | 0.27 | 9.8 |
| Style Emb. (Wegmann et al., 2022) | 30.9 | 0.17 | 0.59 | 0.50 | 1.07 | 0.29 | 6.9 |
| StyleDistance (Patel et al., 2025) | 33.4 | 0.17 | 0.56 | 0.50 | 1.03 | 0.23 | 7.1 |
| StyleDistance (synth.) | 28.1 | 0.15 | 0.54 | 0.50 | 1.06 | 0.12 | 5.5 |

Table 12: Style-encoder comparison on the held-out 50-author Gutenberg test set; **Bold** = best, <u>underlined</u> = second best.

resulting in representations that rely on authorial style rather than content.

**Training hyperparameters.** We use AdamW (default in `sentence-transformers`) as our optimizer with a learning rate of $2 \times 10^{-5}$ and a warmup ratio of 0.1. We train the model with a batch size of 32 and for 20 epochs using one fixed seed (42).

**Encoder performance.** We evaluate the trained encoder on the held-out 50-author Gutenberg test set under the same closed-set authorship-identification and same-author / different-author verification protocols used to compare style encoders (Table 12). On 50-way closed-set attribution — R@$n$ is the share of chunks whose correct author is ranked in the top $n$, and MRR the mean reciprocal rank of that author — the encoder attains R@1 = 18.1% (random baseline 2.0%), R@8 = 58.3% and MRR = 0.32. On same-author vs. different-author verification over 5,000 pairs (2,500 same-author, 2,500 different-author) it reaches AUC = 0.72 and C@1 = 0.64. Embedding-space geometry shows clear separation between authors: the ratio of mean intra-author to mean inter-author cosine similarity (**sep.**) is 7.18, with a corresponding Cohen's $d$ of 0.82, confirming that the encoder places same-author chunks substantially closer than different-author chunks despite the topic-controlled training regime. Under the metrics standard in this literature (Rivera-Soto et al., 2021; Patel et al., 2024) — Recall@8, MRR, and verification AUC — CAV is close to STAR (63.7%, 0.38, 0.71) and ahead of every off-the-shelf encoder we tested, with the largest separation ratio of any encoder in the comparison — making it the strongest encoder we can hold out of the optimization, since STAR itself provides the training reward.

Note, that we use this encoder solely as a **held-out re-evaluator**. It is never queried during adapter training, related-author discovery, weight optimization, or any selection step in AuthorMix. Its purpose is to test whether the relative ranking of methods is robust to the choice of style encoder, complementing the STAR-based primary evaluation (Appendix C.1).

# C Results & Analysis

## C.1 Full Results Table

Table 13 reports the complete results and all AuthorMix configurations across $k \in \{1, \ldots, 10\}$. The right side of the table reports style transfer accuracy and meaning preservation under scoring functions held out from the AuthorMix optimization pipeline (a CAV encoder (Wegmann et al., 2022) trained from scratch on Gutenberg literary text, see Appendix B.5.3) paired with SBERT-based meaning similarity (Reimers and Gurevych, 2019). We can see that under the held-out scoring functions, the relative ordering of methods is preserved, with AuthorMix GRPO-layer-wise achieving the best $\text{Joint}_{\text{CAV}}$ score (0.50 at best $k = 4$, vs. 0.45 for ASTRAPOP-DPO, 0.40 for TinyStyler, 0.43–0.45 for STYLL variants), and the layer-wise variant continuing to outperform the adapter-wise variant. STYLL retains the highest $\text{Toward}_{\text{CAV}}$ score (0.37) but at the cost of a low SBERT score (0.65), mirroring the style and meaning trade-off observed under STAR/MIS. The agreement between two style encoders trained with different objectives and architectures, and two non-overlapping meaning-preservation metrics, indicates that AuthorMix's advantage is not an artefact of the STAR/MIS-based evaluation pipeline. Appendix C.2 tests these differences pairwise over the 100 source–target pairs.

## C.2 Significance of the Automatic-Metric Differences

Table 14 reports Wilcoxon signed-rank tests and paired effect sizes ($d_z$) comparing AuthorMix (GRPO-LW, $k = 4$) against each style-transfer baseline, computed on the per-pair scores under the primary scoring functions.

| Method | k | *Primary scoring functions (STAR + MIS)* | | | | | *Held-out scoring functions (CAV + SBERT)* | | |
|---|---|---|---|---|---|---|---|---|---|
| | | **Toward↑** | **MIS↑** | **Joint↑** | **CoLA↑** | **Away↑** | **Toward (CAV)↑** | **SBERT↑** | **Joint (CAV)↑** |
| Neutral Text† | — | 0.01 | 0.79 | 0.05 | **0.89** | **0.91** | 0.21 | 0.76 | 0.39 |
| Zero-shot (LLaMA-8B) | — | 0.07 | 0.81 | 0.17 | 0.88 | 0.81 | 0.23 | 0.81 | 0.40 |
| Few-shot (GPT-5.1) | — | 0.08 | 0.81 | 0.20 | 0.72 | 0.65 | 0.23 | 0.78 | 0.39 |
| STYLL (GPT-3.5) | — | 0.07 | 0.68 | 0.16 | 0.88 | 0.86 | **0.37** | 0.65 | 0.45 |
| STYLL (GPT-5.1) | — | 0.15 | 0.62 | 0.27 | 0.71 | 0.71 | 0.31 | 0.76 | 0.43 |
| ASTRAPOP-SFT | — | 0.16 | 0.63 | 0.29 | 0.80 | 0.63 | 0.26 | 0.73 | 0.37 |
| ASTRAPOP-DPO | — | **0.17** | 0.63 | 0.29 | 0.83 | 0.78 | 0.36 | 0.71 | 0.45 |
| TinyStyler | — | 0.16 | 0.75 | 0.31 | 0.68 | 0.66 | 0.26 | 0.79 | 0.40 |
| TinyStyler-Sim | — | 0.16 | 0.80 | 0.33 | 0.68 | 0.63 | 0.24 | 0.82 | 0.39 |
| Single adapter | 1 | 0.06 | 0.86 | 0.17 | 0.73 | 0.46 | 0.13 | **0.89** | 0.29 |
| *AuthorMix — LoRAHub, adapter-wise* | | | | | | | | | |
| | 2 | 0.10 | 0.84 | 0.23 | 0.86 | 0.74 | 0.30 | 0.85 | 0.45 |
| | 3 | 0.11 | 0.85 | 0.25 | 0.85 | 0.71 | 0.28 | 0.87 | 0.44 |
| | 4 | 0.10 | 0.84 | 0.24 | 0.85 | 0.73 | 0.28 | 0.86 | 0.44 |
| | 5 | 0.11 | 0.84 | 0.25 | 0.84 | 0.70 | 0.28 | 0.86 | 0.44 |
| | 6 | 0.11 | 0.85 | 0.25 | 0.83 | 0.67 | 0.26 | 0.86 | 0.41 |
| | 7 | 0.11 | 0.86 | 0.24 | 0.81 | 0.61 | 0.23 | 0.88 | 0.38 |
| | 8 | 0.10 | 0.84 | 0.24 | 0.84 | 0.69 | 0.26 | 0.86 | 0.42 |
| | 9 | 0.11 | 0.86 | 0.25 | 0.82 | 0.62 | 0.23 | 0.88 | 0.39 |
| | 10 | 0.10 | 0.85 | 0.24 | 0.81 | 0.63 | 0.22 | 0.87 | 0.38 |
| *AuthorMix — LoRAHub, layerwise* | | | | | | | | | |
| | 2 | 0.11 | 0.86 | 0.26 | 0.83 | 0.65 | 0.27 | **0.89** | 0.43 |
| | 3 | 0.12 | 0.86 | 0.27 | 0.83 | 0.64 | 0.27 | **0.89** | 0.42 |
| | 4 | 0.12 | 0.86 | 0.27 | 0.83 | 0.65 | 0.27 | **0.89** | 0.42 |
| | 5 | 0.12 | 0.86 | 0.26 | 0.83 | 0.65 | 0.27 | **0.89** | 0.43 |
| | 6 | 0.12 | **0.87** | 0.27 | 0.83 | 0.64 | 0.26 | **0.89** | 0.42 |
| | 7 | 0.12 | 0.86 | 0.27 | 0.83 | 0.64 | 0.26 | **0.89** | 0.42 |
| | 8 | 0.12 | **0.87** | 0.27 | 0.83 | 0.64 | 0.26 | **0.89** | 0.42 |
| | 9 | 0.12 | **0.87** | 0.27 | 0.83 | 0.64 | 0.26 | **0.89** | 0.42 |
| | 10 | 0.12 | **0.87** | 0.27 | 0.83 | 0.64 | 0.26 | **0.89** | 0.42 |
| *AuthorMix — GRPO, adapter-wise* | | | | | | | | | |
| | 2 | 0.12 | 0.79 | 0.27 | 0.87 | 0.75 | 0.32 | 0.80 | 0.46 |
| | 3 | 0.13 | 0.78 | 0.29 | 0.86 | 0.72 | 0.31 | 0.81 | 0.45 |
| | 4 | 0.12 | 0.77 | 0.26 | 0.85 | 0.73 | 0.29 | 0.81 | 0.44 |
| | 5 | 0.11 | 0.75 | 0.25 | 0.86 | 0.77 | 0.31 | 0.80 | 0.45 |
| | 6 | 0.13 | 0.78 | 0.28 | 0.88 | 0.75 | 0.33 | 0.79 | 0.47 |
| | 7 | 0.13 | 0.78 | 0.29 | 0.87 | 0.73 | 0.32 | 0.80 | 0.46 |
| | 8 | 0.12 | 0.76 | 0.27 | 0.86 | 0.75 | 0.30 | 0.80 | 0.45 |
| | 9 | 0.12 | 0.76 | 0.27 | 0.86 | 0.76 | 0.31 | 0.80 | 0.45 |
| | 10 | 0.12 | 0.73 | 0.26 | 0.86 | 0.77 | 0.31 | 0.80 | 0.45 |
| *AuthorMix — GRPO, layerwise* | | | | | | | | | |
| | 2 | 0.15 | 0.83 | 0.32 | 0.85 | 0.69 | 0.32 | 0.83 | 0.48 |
| | 3 | 0.14 | 0.83 | 0.31 | 0.85 | 0.71 | 0.33 | 0.82 | 0.48 |
| | 4 | 0.16 | 0.83 | **0.34** | 0.83 | 0.67 | 0.35 | 0.84 | 0.50 |
| | 5 | 0.15 | 0.82 | 0.32 | 0.84 | 0.72 | 0.32 | 0.82 | 0.48 |
| | 6 | 0.15 | 0.82 | 0.33 | 0.83 | 0.71 | **0.37** | 0.83 | **0.51** |
| | 7 | 0.11 | 0.80 | 0.26 | 0.83 | 0.77 | 0.35 | 0.80 | 0.50 |
| | 8 | 0.12 | 0.82 | 0.28 | 0.84 | 0.76 | 0.35 | 0.81 | 0.49 |
| | 9 | 0.13 | 0.80 | 0.30 | 0.83 | 0.77 | 0.35 | 0.80 | 0.50 |
| | 10 | 0.11 | 0.79 | 0.26 | 0.82 | 0.80 | 0.33 | 0.80 | 0.47 |

Table 13: Full results for all baselines and all AuthorMix configurations across $k \in \{2, ..., 10\}$ related authors. All scores averaged over 100 source–target pairs × 10 evaluation runs. The left side reports our primary STAR-based evaluation; the right side reports the held-out scoring functions: the CAV style encoder (Wegmann et al., 2022) with SBERT semantic similarity (Reimers and Gurevych, 2019). **Bold** = best within each block; underline = second-best. †Reference bound.

AuthorMix attains a significantly higher Joint score than every baseline, including the strongest one (TinyStyler, $p = 0.03$), and a significantly higher MIS than all of them, with large effect sizes against the trained baselines ($d_z = 1.66$–$2.44$). On Toward, the differences against TinyStyler and the two ASTRAPOP variants are small and not significant ($p \geq 0.18$), while STYLL and few-shot GPT-5.1 are significantly below AuthorMix. AuthorMix is therefore best or statistically tied on each individual axis under the primary scoring functions, and strictly best on the combined score.

| Baseline | Toward | | MIS | | Joint | |
|---|---|---|---|---|---|---|
| | $d_z$ | $p$ | $d_z$ | $p$ | $d_z$ | $p$ |
| TinyStyler | −0.04 | n.s. | 1.66 | <0.001 | 0.22 | 0.03 |
| ASTRAPOP-DPO | −0.10 | n.s. | 2.44 | <0.001 | 0.41 | <0.001 |
| ASTRAPOP-SFT | −0.09 | n.s. | 2.04 | <0.001 | 0.58 | <0.001 |
| STYLL | 0.98 | <0.001 | 1.97 | <0.001 | 1.41 | <0.001 |
| Few-shot | 0.77 | <0.001 | 0.35 | 0.003 | 0.83 | <0.001 |

Table 14: Paired comparisons of AuthorMix (GRPO-LW, $k = 4$) against each baseline. $d_z$ is the paired effect size (positive favours AuthorMix); $p$ is from a Wilcoxon signed-rank test. n.s. = not significant at $\alpha = 0.05$.

### C.3 Cross-Base-Model Results: Qwen2.5-7B-Instruct

To assess whether AuthorMix's gains depend on the choice of base model, we re-train the entire pipeline using Qwen2.5-7B-Instruct[10] as the underlying language model. The high-resource adapter library is re-trained on the same 10 authors as the LLaMA pipeline (Appendix B.3.1), the same neutralization prompt, the same evaluation protocol, and the same 100 source–target pairs. For a fair comparison, ASTRAPOP-SFT and ASTRAPOP-DPO are also re-trained with Qwen2.5-7B-Instruct as their base model. Baselines that do not depend on the AuthorMix backbone (Neutral Text, Few-shot, STYLL, TinyStyler) carry over from Table 13 unchanged.

Table 15 reports the full sweep across $k \in \{1, ..., 10\}$ for all four AuthorMix variants. Following the convention used for the LLaMA results (Section 5), we select the best-performing $k$ for each variant based on the primary Joint score (the optimization metric), breaking ties towards the smallest $k$, and report the corresponding held-out CAV+SBERT scores at that same $k$.

**What carries over.** The principal result holds on Qwen2.5-7B-Instruct: GRPO layer-wise remains the strongest AuthorMix variant (best Joint 0.31 at $k = 7$; the others peak at 0.23 for GRPO-AW, 0.21 for LoRAHub-AW and 0.00 for LoRAHub-LW), still beats adapter-wise GRPO mixing, and still outperforms both re-trained ASTRAPOP variants on Joint (0.31 vs. 0.28 and 0.27) and on meaning preservation (MIS = 0.79 vs. 0.67 and 0.64). Under the held-out CAV+SBERT scoring functions at the same $k$ it scores higher than every baseline ($\text{Joint}_{\text{CAV}}$ = 0.48, vs. 0.44 for ASTRAPOP-DPO, 0.40 for TinyStyler and 0.39 for TinyStyler-Sim), with higher semantic similarity as well (SBERT = 0.80 vs. 0.73). The absolute Joint score is slightly lower than under LLaMA-3.1-8B ($0.34 \rightarrow 0.31$), consistent with the smaller backbone.

**What does not.** The complete ranking of variants changes. On the primary metric GRPO-LW matches TinyStyler (0.31) but trails TinyStyler-Sim (0.33), which it had surpassed on LLaMA. LoRAHub layer-wise does not merely underperform, as on LLaMA, but collapses: NGOpt converges to near-empty outputs across the full sweep $k \in \{2, ..., 10\}$ (MIS $\approx$ 0.03–0.13, Toward $\approx 0$, with a trivially perfect Away $\approx 1.0$), consistent with our LLaMA observation (Appendix C.5.1) that gradient-free search struggles in the substantially larger layer-wise search space. The two LoRAHub variants therefore swap places, and adapter-wise becomes the better gradient-free option (best $k = 3$, $\text{Joint}_{\text{CAV}}$ = 0.46). We claim cross-backbone robustness for this principal comparison, not for the full ordering of all eight rows.

### C.4 Primary Metric Distributions

#### C.4.1 MIS Score Distribution

Figure 5 and Figure 6 show the full MIS score distributions across all 100 source–target pairs, including TinyStyler-Sim post-publication variant. The KDE plot (Figure 6) annotates each method with its mean ($\mu$) and standard deviation ($\sigma$).

AuthorMix achieves the highest mean MIS ($\mu = 0.83$) with the lowest variance ($\sigma = 0.059$) among all style-transfer methods — nearly half the standard deviation of ASTRAPOP-SFT ($\sigma = 0.115$) and ASTRAPOP-DPO ($\sigma = 0.107$). This confirms that AuthorMix not only preserves mean-ing better on average but does so *consistently* across source–target pairs.

[10] Qwen/Qwen2.5-7B-Instruct

| Method | k | *Primary scoring functions (STAR + MIS)* | | | | | *Held-out scoring functions (CAV + SBERT)* | | |
|---|---|---|---|---|---|---|---|---|---|
| | | **Toward↑** | **MIS↑** | **Joint↑** | **CoLA↑** | **Away↑** | **Toward$_{CAV}$↑** | **SBERT↑** | **Joint$_{CAV}$↑** |
| Neutral Text† | — | 0.01 | 0.79 | 0.05 | **0.89** | 0.91 | 0.21 | 0.76 | 0.39 |
| Zero-shot (LLaMA-8B) | — | 0.07 | 0.81 | 0.17 | <u>0.88</u> | 0.81 | 0.23 | 0.81 | 0.40 |
| Few-shot (GPT-5.1) | — | 0.08 | 0.81 | 0.20 | 0.72 | 0.65 | 0.23 | 0.78 | 0.39 |
| STYLL (GPT-3.5) | — | 0.07 | 0.68 | 0.16 | <u>0.88</u> | 0.86 | **0.37** | 0.65 | 0.45 |
| STYLL (GPT-5.1) | — | 0.15 | 0.62 | 0.27 | 0.71 | 0.71 | 0.31 | 0.76 | 0.43 |
| TinyStyler | — | **0.16** | 0.75 | <u>0.31</u> | 0.68 | 0.66 | 0.26 | 0.79 | 0.40 |
| TinyStyler-Sim | — | **0.16** | 0.80 | **0.33** | 0.68 | 0.63 | 0.24 | 0.82 | 0.39 |
| ASTRAPOP-SFT (Qwen2.5) | — | 0.14 | 0.64 | 0.27 | 0.80 | 0.62 | 0.22 | 0.73 | 0.34 |
| ASTRAPOP-DPO (Qwen2.5) | — | <u>0.15</u> | 0.67 | 0.28 | 0.85 | 0.79 | **0.34** | 0.73 | 0.44 |
| Single adapter | 1 | 0.05 | **0.84** | 0.14 | 0.71 | 0.54 | 0.19 | **0.85** | 0.35 |
| *AuthorMix (Qwen2.5) — LoRAHub, adapter-wise* | | | | | | | | | |
| | 2 | 0.08 | 0.82 | 0.20 | 0.80 | 0.73 | **0.34** | 0.81 | <u>0.49</u> |
| | 3 | 0.08 | **0.84** | 0.21 | 0.79 | 0.67 | 0.30 | 0.82 | 0.46 |
| | 4 | 0.08 | <u>0.83</u> | 0.20 | 0.80 | 0.71 | 0.33 | 0.81 | <u>0.49</u> |
| | 5 | 0.08 | **0.84** | 0.20 | 0.78 | 0.65 | 0.30 | 0.82 | 0.44 |
| | 6 | 0.07 | **0.84** | 0.19 | 0.79 | 0.66 | 0.29 | 0.82 | 0.44 |
| | 7 | 0.06 | <u>0.83</u> | 0.17 | 0.80 | 0.71 | 0.31 | 0.81 | 0.45 |
| | 8 | 0.08 | 0.82 | 0.19 | 0.81 | 0.74 | **0.34** | 0.80 | **0.50** |
| | 9 | 0.09 | <u>0.83</u> | 0.20 | 0.80 | 0.71 | **0.34** | 0.81 | <u>0.49</u> |
| | 10 | 0.08 | **0.84** | 0.20 | 0.78 | 0.62 | 0.27 | <u>0.83</u> | 0.42 |
| *AuthorMix (Qwen2.5) — LoRAHub, layerwise‡* | | | | | | | | | |
| | 2 | 0.00 | 0.09 | 0.00 | 0.41 | **1.00** | 0.20 | 0.19 | 0.16 |
| | 3 | 0.00 | 0.06 | 0.00 | 0.52 | **1.00** | 0.20 | 0.10 | 0.12 |
| | 4 | 0.00 | 0.05 | 0.00 | 0.53 | **1.00** | 0.19 | 0.11 | 0.12 |
| | 5 | 0.00 | 0.13 | 0.00 | 0.55 | <u>0.99</u> | 0.20 | 0.17 | 0.14 |
| | 6 | 0.00 | 0.05 | 0.00 | 0.53 | **1.00** | 0.19 | 0.11 | 0.12 |
| | 7 | 0.00 | 0.04 | 0.00 | 0.53 | **1.00** | 0.18 | 0.10 | 0.11 |
| | 8 | 0.00 | 0.03 | 0.00 | 0.53 | **1.00** | 0.14 | 0.10 | 0.10 |
| | 9 | 0.00 | 0.03 | 0.00 | 0.53 | **1.00** | 0.14 | 0.10 | 0.09 |
| | 10 | 0.00 | 0.03 | 0.00 | 0.53 | **1.00** | 0.14 | 0.10 | 0.10 |
| *AuthorMix (Qwen2.5) — GRPO, adapter-wise* | | | | | | | | | |
| | 2 | 0.08 | 0.80 | 0.22 | 0.73 | 0.67 | 0.27 | 0.80 | 0.44 |
| | 3 | 0.08 | 0.78 | 0.22 | 0.70 | 0.66 | 0.29 | 0.78 | 0.44 |
| | 4 | 0.08 | 0.79 | 0.22 | 0.72 | 0.65 | 0.26 | 0.80 | 0.41 |
| | 5 | 0.07 | 0.76 | 0.19 | 0.70 | 0.71 | 0.30 | 0.77 | 0.45 |
| | 6 | 0.08 | 0.79 | 0.22 | 0.70 | 0.71 | 0.30 | 0.78 | 0.45 |
| | 7 | 0.07 | 0.79 | 0.18 | 0.71 | 0.73 | 0.29 | 0.78 | 0.44 |
| | 8 | 0.07 | 0.80 | 0.20 | 0.70 | 0.73 | 0.29 | 0.78 | 0.45 |
| | 9 | 0.08 | 0.79 | 0.22 | 0.71 | 0.72 | 0.30 | 0.79 | 0.45 |
| | 10 | 0.10 | 0.70 | 0.23 | 0.69 | 0.69 | 0.28 | 0.73 | 0.41 |
| *AuthorMix (Qwen2.5) — GRPO, layerwise* | | | | | | | | | |
| | 2 | 0.11 | 0.83 | 0.27 | 0.75 | 0.64 | 0.28 | 0.81 | 0.44 |
| | 3 | 0.13 | 0.81 | 0.30 | 0.77 | 0.63 | 0.28 | 0.81 | 0.44 |
| | 4 | 0.12 | 0.80 | 0.27 | 0.74 | 0.70 | 0.33 | 0.80 | 0.48 |
| | 5 | 0.13 | 0.80 | 0.29 | 0.78 | 0.69 | 0.31 | 0.80 | 0.47 |
| | 6 | 0.13 | 0.79 | 0.30 | 0.79 | 0.67 | 0.30 | 0.80 | 0.45 |
| | 7 | 0.14 | 0.79 | <u>0.31</u> | 0.78 | 0.65 | 0.33 | 0.80 | 0.48 |
| | 8 | 0.13 | 0.77 | 0.29 | 0.80 | 0.73 | 0.31 | 0.79 | 0.46 |
| | 9 | 0.14 | 0.75 | 0.30 | 0.79 | 0.71 | 0.30 | 0.77 | 0.45 |
| | 10 | 0.12 | 0.78 | 0.27 | 0.77 | 0.73 | 0.32 | 0.77 | 0.46 |

Table 15: **Qwen2.5-7B-Instruct results across all AuthorMix configurations.** Format mirrors Table 13. Baselines (Neutral, Few-shot, STYLL, TinyStyler) carry over from the LLaMA evaluation unchanged since they do not depend on the AuthorMix backbone; ASTRAPOP-SFT/DPO are re-trained with Qwen2.5-7B-Instruct as the base model for fair comparison. **Bold** = best; <u>underline</u> = second-best.

TinyStyler-Sim ($\mu = 0.80$, $\sigma = 0.072$) improves over TinyStyler ($\mu = 0.75$, $\sigma = 0.071$) by adding further candidate generation and filtering candidates for semantic similarity, narrowing the gap to AuthorMix. However, AuthorMix still achieves a +4% higher mean with lower variance, and without requiring any candidate filtering step.

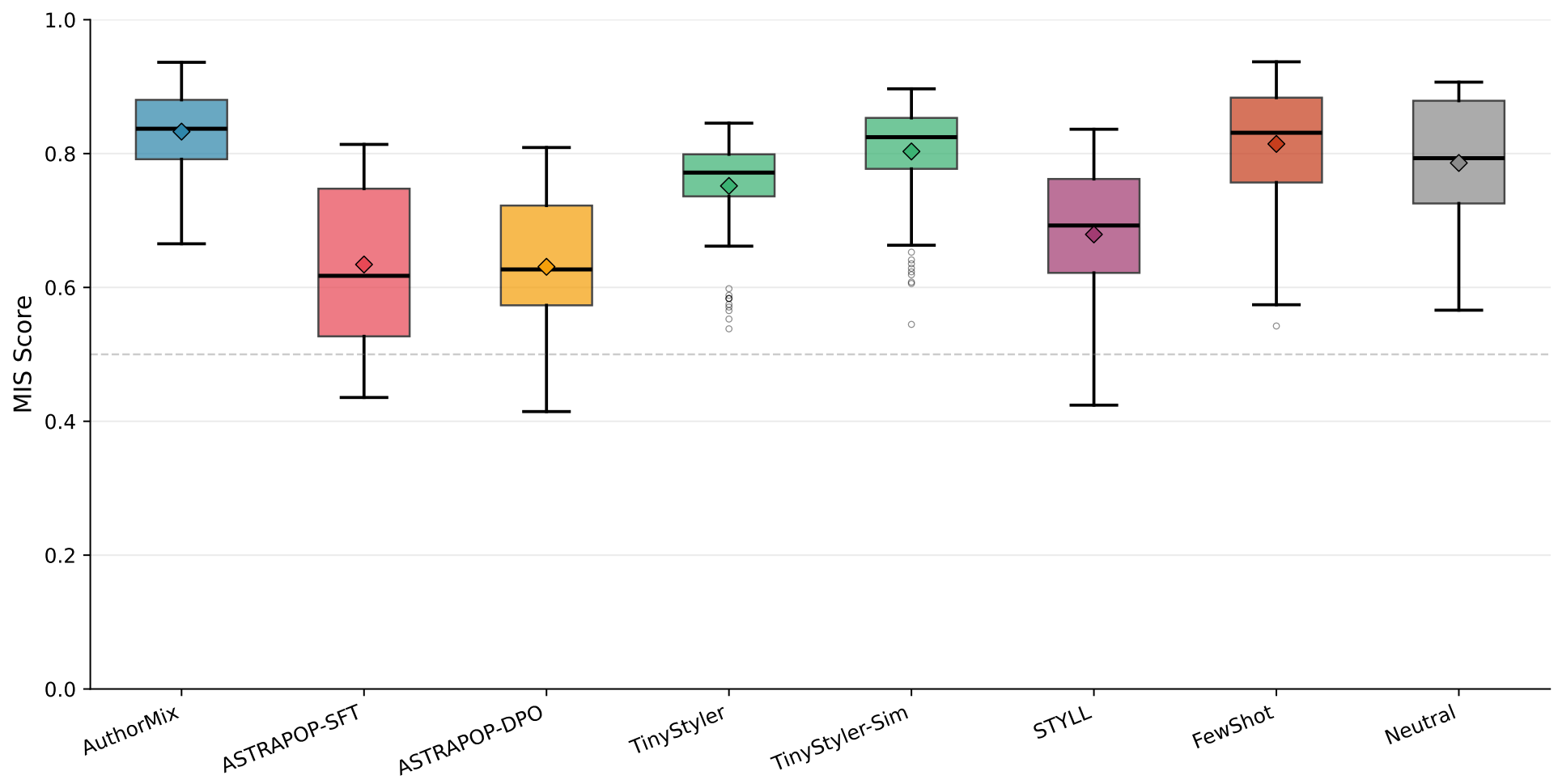

Figure 5: Box plot of MIS score distributions across all 100 source–target pairs for each method (including TinyStyler-Sim). Boxes show the interquartile range; diamonds mark the mean; whiskers extend to 1.5×IQR; circles are outliers. The dashed line at 0.5 indicates the threshold below which semantic equivalence breaks down.

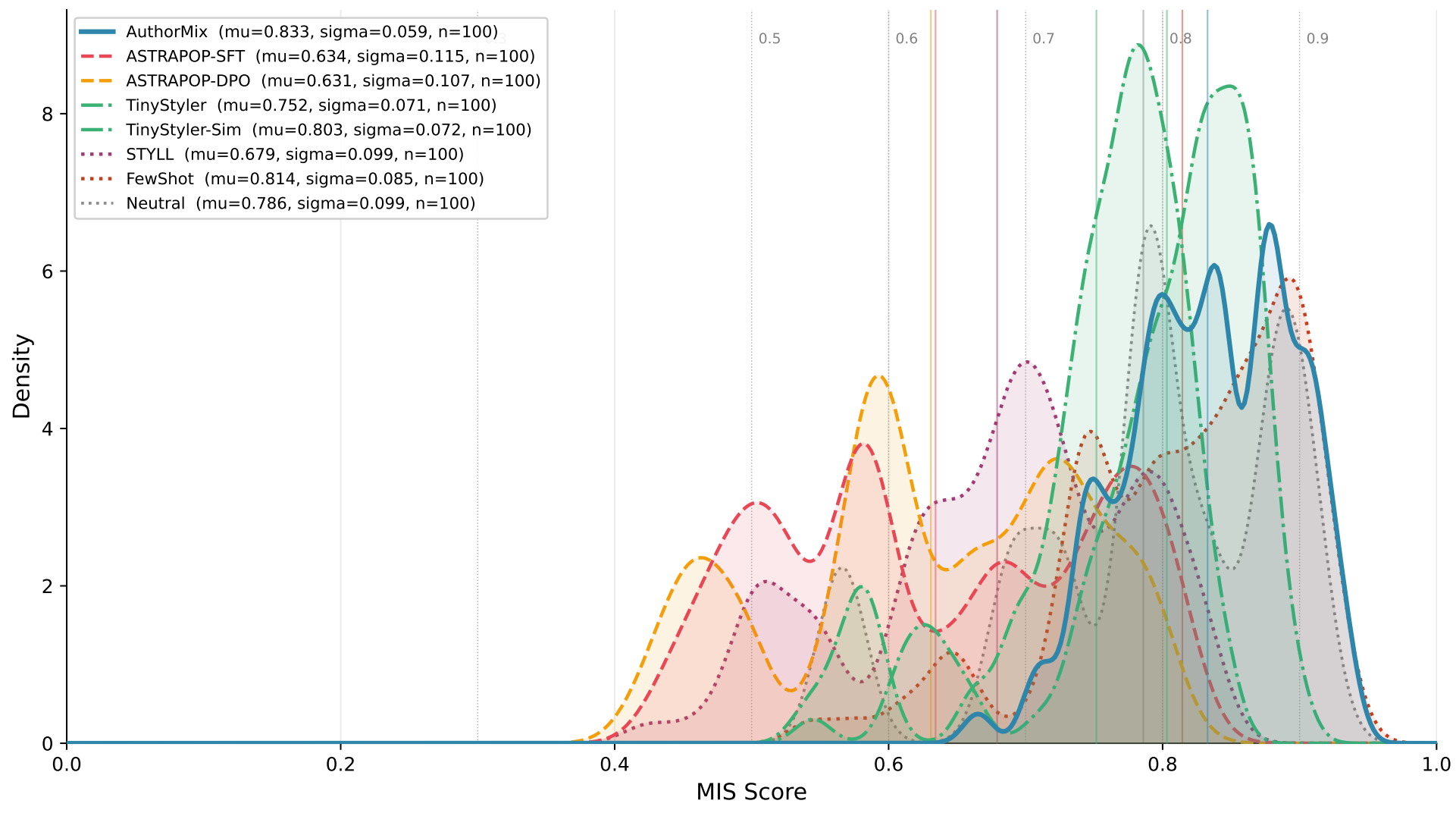

Figure 6: Kernel density estimate (KDE) of MIS score distributions for all methods. Legend shows the mean ($\mu$), standard deviation ($\sigma$), and number of evaluation pairs ($n = 100$) for each method. Vertical lines mark the mean of each distribution.

ASTRAPOP variants show the widest spread, with density mass extending well below the 0.5 threshold (dashed line in Figure 5), confirming that aggressive style transfer via DPO/SFT frequently comes at the cost of meaning. STYLL ($\mu = 0.68$, $\sigma = 0.099$) occupies a middle ground, while FewShot ($\mu = 0.81$) and Neutral ($\mu = 0.79$) score high on MIS but apply minimal style transfer.

### C.4.2 Toward and Joint Distributions

We repeat the distributional analysis applied to MIS (above) for the Toward and Joint primary metrics, to show the variability for all primary metrics rather than only MIS. The Toward distribution (Figure 7, Figure 8) shows that the four style-transferring methods — AuthorMix, ASTRAPOP-SFT, ASTRAPOP-DPO, TinyStyler — cluster around the same mean ($\mu \approx 0.16$), confirming that AuthorMix matches the strongest baselines on absolute style transfer strength. STYLL and FewShot achieve a noticeably lower Toward score ($\mu \approx 0.07$–$0.08$), consistent with the main-table numbers. As expected for a baseline that does not perform style transfer, neutral text ($\mu = 0.01$) remains close to zero.

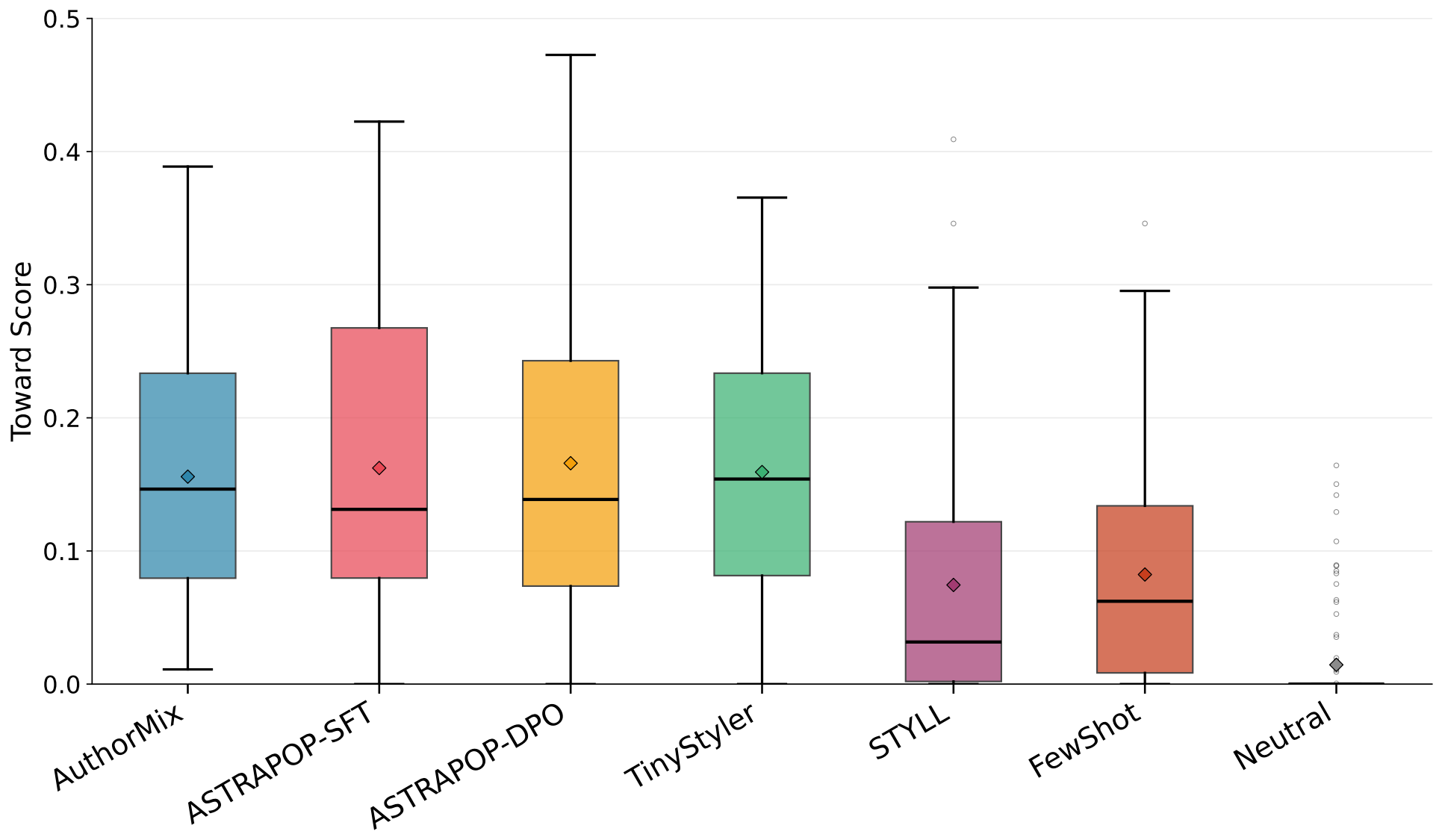

Figure 7: Toward-score distribution across the 100 source–target pairs for each method (Patel-scaled toward; same generations as Figure 5). Boxes show the interquartile range; diamonds mark the mean; whiskers extend to 1.5× IQR. Neutral sits at $\mu \approx 0.01$ — as expected for a non-style-transferring reference.

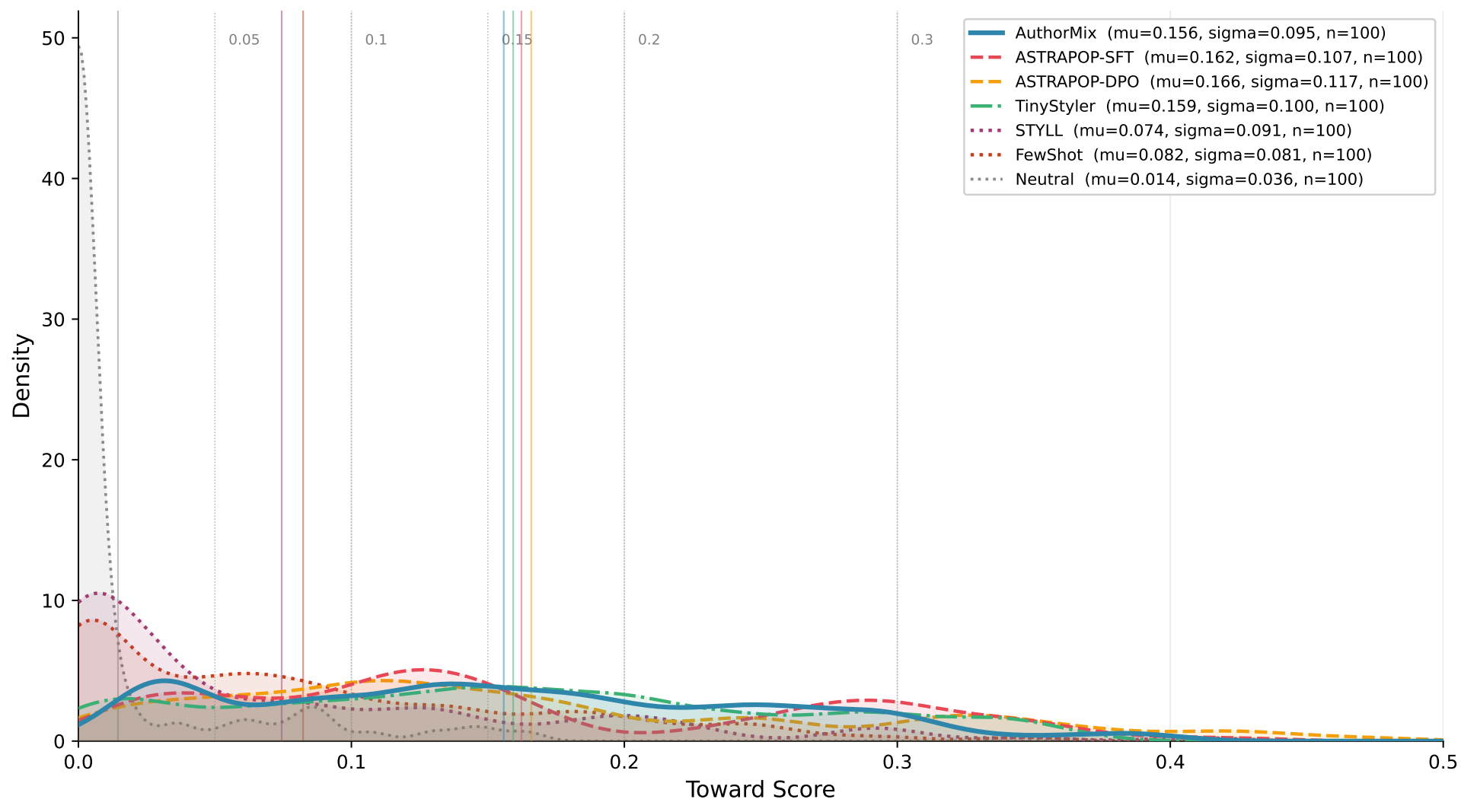

Figure 8: Kernel density estimate (KDE) of Toward-score distributions for all methods. Legend shows mean ($\mu$), standard deviation ($\sigma$), and number of pairs ($n = 100$) per method.

The Joint score distribution (Figure 9, Figure 10) makes the decisive separation visible: AuthorMix has both the highest mean Joint score ($\mu = 0.34$) and a tighter distribution than the ASTRAPOP variants or STYLL. The advantage over ASTRAPOP-SFT/DPO ($\mu \approx 0.29$) is exactly the style–meaning trade-off seen in the Table 13 both methods transfer style with similar strength, but ASTRAPOP loses substantially more meaning, and the Joint geometric mean penalizes that loss. TinyStyler ($\mu = 0.31$) is the closest baseline on Joint mean but with a wider lower tail, while FewShot and STYLL fall further behind due to their lower Toward.

### C.5 Gradient-Based vs. Gradient-Free Weight Learning

#### C.5.1 Layer Weight Analysis

Figure 12 shows the per-layer adapter weights learned by GRPO for a single target author (Robert W. Chambers, $k$=4). The weights are far from uniform: adapter PGW (P. G. Wodehouse) receives strong positive weights throughout, while adapters VW (Virginia Woolf) and GO (George Orwell) are actively suppressed with negative weights (down to $-1.0$) at deep layers 24–31 — the layers that tend to encode discourse-level and syntactic patterns (Clark et al., 2019). The early layers (0–10) show moderate mixed-sign weights, and the middle layers (14–21) are dominated by strong positive weights for PGW and VW. This is principled: following the task-arithmetic framework (Ilharco et al., 2023) and TIES-Merging (Yadav et al., 2023), LoRA adapters act as task vectors in parameter space, and subtracting a task vector steers the model *away* from the style that adapter encodes.

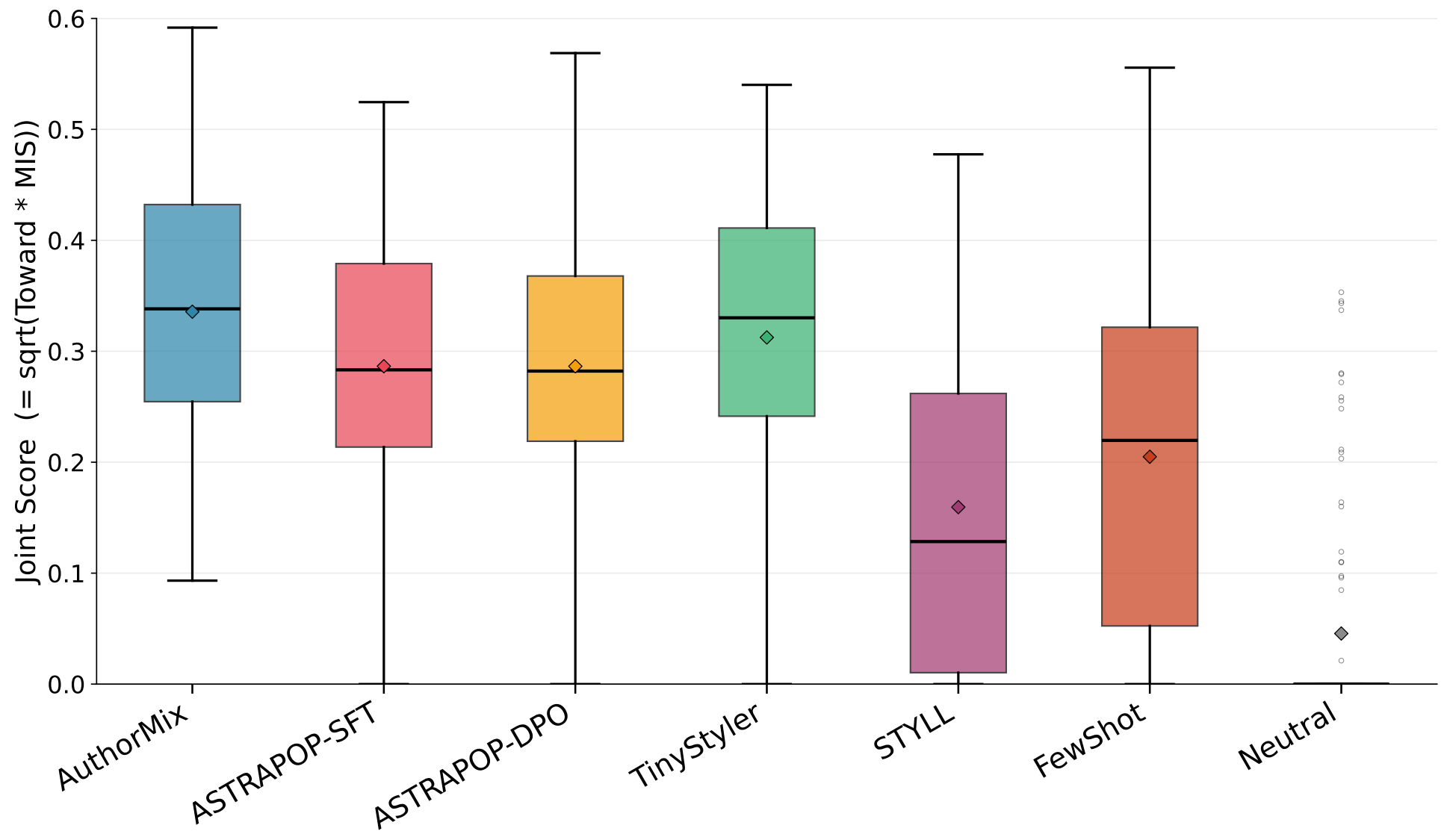

Figure 9: Joint-score distribution (Joint $= \sqrt{\text{Toward} \times \text{MIS}}$) across the 100 source–target pairs. AuthorMix has the highest mean **and** the tightest box, confirming that its lead on the Joint metric is consistent across source–target pairs rather than driven by a few outliers.

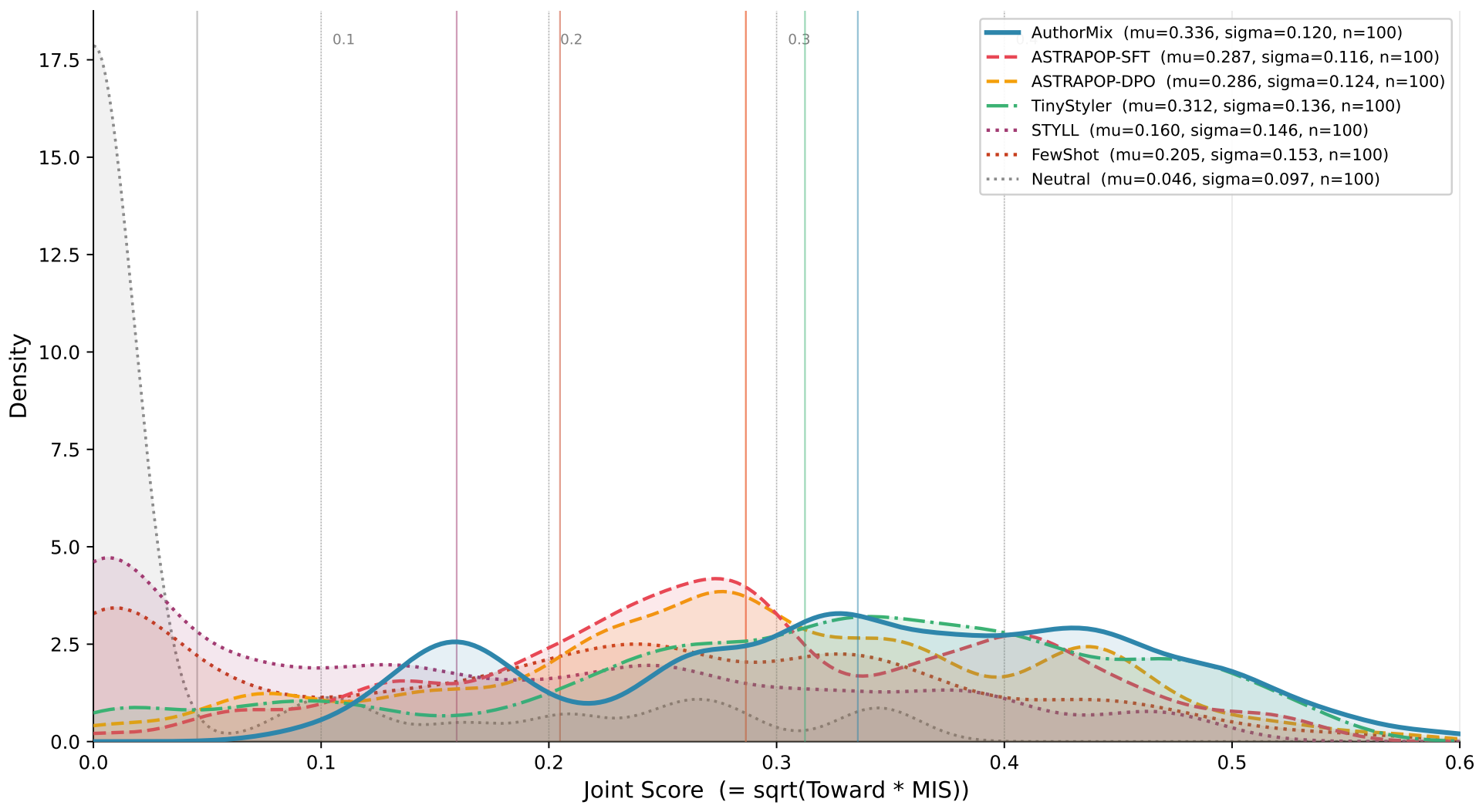

Figure 10: Kernel density estimate (KDE) of Joint-score distributions. Generated by the same pipeline that produces Figure 6 and Figure 8).

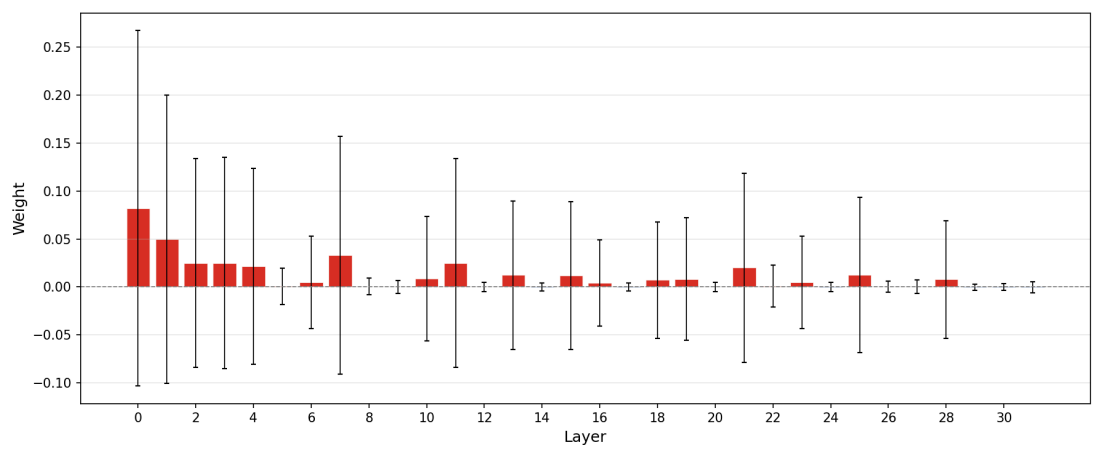

Figure 11: LoRAHub layerwise: global average of learned layer weights across all 10 target authors and all $k$ values. All weights are near zero, indicating that gradient-free optimization does not produce structured layerwise patterns. Vertical lines indicate $\pm 1$ standard deviation.

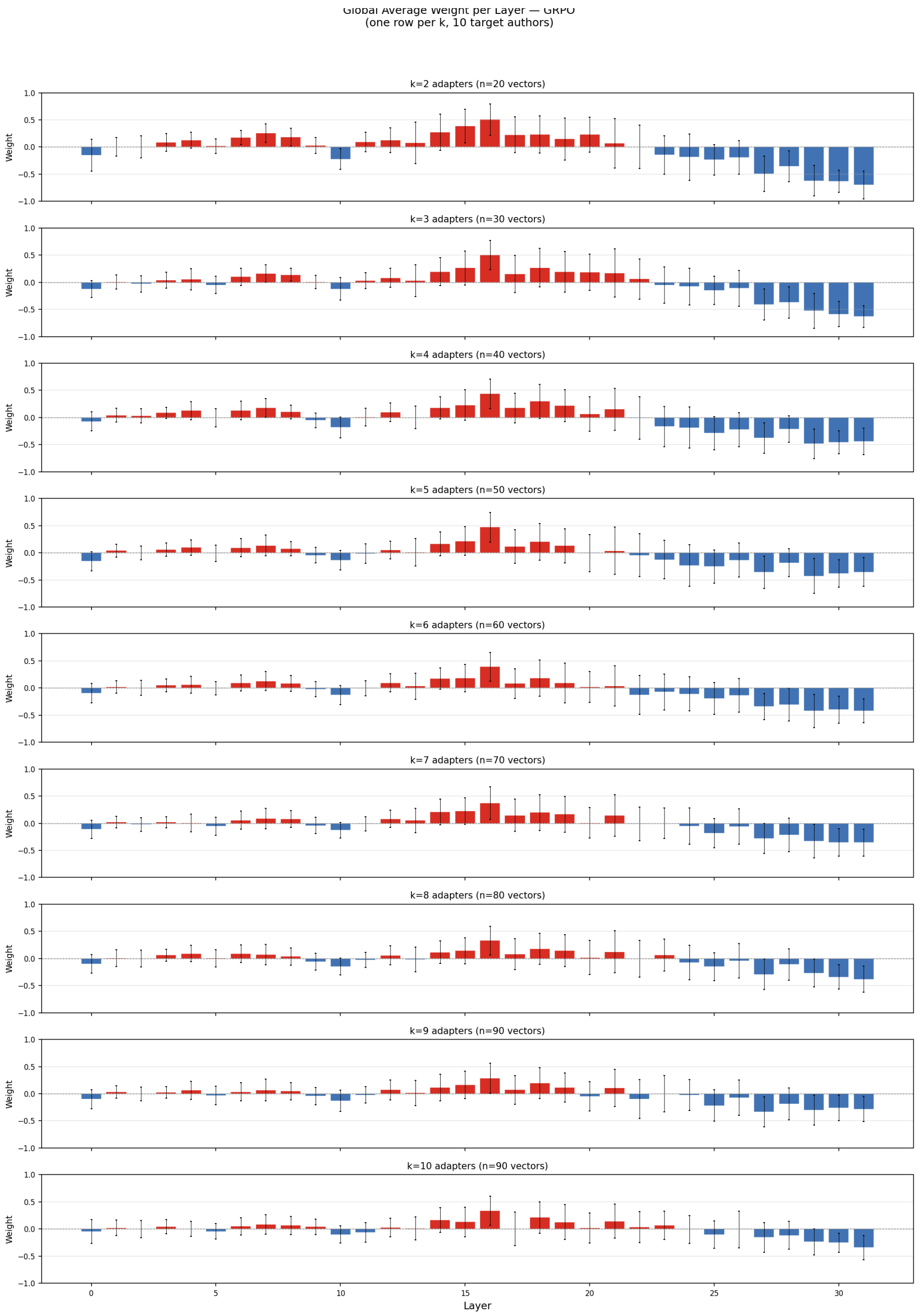

Figure 13: GRPO layerwise: global average of learned layer weights per $k$ ($k$=2–10), averaged across all 10 target authors. Each subplot shows one $k$ value. Red = positive weight; blue = negative weight. Vertical lines indicate $\pm$ 1 standard deviation across target authors.

Figure 13 breaks down the GRPO layerwise weights by $k$ (number of adapters). The pattern described in the main text — positive middle layers, negative deep layers — is consistent across all $k$ values ($k$=2–10), confirming that this structure is a robust property of GRPO optimization rather than an artifact of a specific $k$.

In contrast, Figure 11 shows the LoRAHub (gradient-free) layerwise weights averaged across all $k$ values and target authors. The weights are near zero across all layers, with no discernible structure — confirming that gradient-free optimization struggles to exploit the layerwise search space effectively.

## C.6 Human Evaluation

We complement the automatic evaluation with a human evaluation study comparing AuthorMix against four strong baselines: ASTRAPOP-DPO, Few-shot GPT-5.1, TinyStyler, and STYLL. We sample 50 source sentences, consisting of 2 sentences for each of 25 source-target author pairs formed from 5 source authors and 5 target authors. Each source sentence is rewritten by all five methods toward the corresponding target author, yielding 250 system outputs. The study was conducted on Prolific, where raters scored three dimensions per item: **Meaning Preservation (MP)** on a 0–2 ordinal scale (2 = fully preserved, 1 = partially preserved, 0 = substantially changed); **Fluency** as a binary judgement (1 = grammatical); and **Style match** as a 2-Alternative Forced Choice (2AFC) in which the participant is shown five example texts from the target author and five from the source author, and asked to identify which set the rewrite's style most resembles (Patel et al., 2024; Liu et al., 2024a). Optionally, participants could also explain their Style match selection. (See materials here Appendix C.13.)

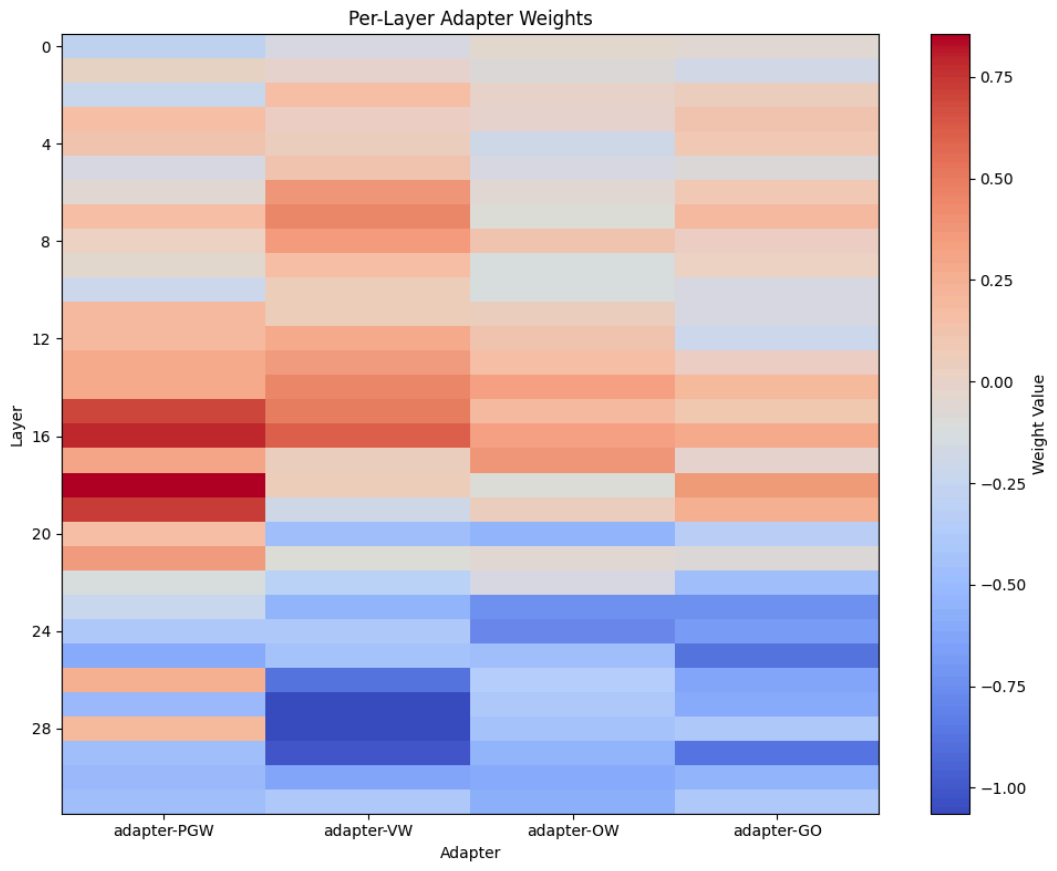

Figure 12: Per-layer adapter weights learned by GRPO for target author Robert W. Chambers ($k$=4). Each column is one of the top-$k$=4 style-similar related authors (PGW, VW, OW, GO); each row is a transformer layer (0–31). Red = positive weight (adapter is drawn upon); blue = negative (adapter is suppressed).

| Dimension | Contrast | $d_z$ | $p_{\text{Holm}}$ |
|---|---|---|---|
| MP | AM − Few-shot | 0.18 | 0.20 |
| MP | AM − STYLL | **0.56** | 0.0005 |
| MP | AM − TinyStyler | **0.75** | <0.0001 |
| MP | AM − ASTRAPOP | **1.11** | <0.0001 |
| Style | AM − Few-shot | **0.52** | 0.002 |
| Style | AM − STYLL | 0.00 | 1.00 |
| Style | AM − TinyStyler | 0.05 | 1.00 |
| Style | AM − ASTRAPOP | 0.13 | 1.00 |
| Fluency | AM − Few-shot | **0.41** | 0.016 |
| Fluency | AM − STYLL | 0.04 | 0.76 |
| Fluency | AM − TinyStyler | **0.94** | <0.0001 |
| Fluency | AM − ASTRAPOP | 0.33 | 0.048 |

Table 16: Per-dimension paired **t**-tests on item-level means (N = 50 paired source sentences, 3 raters per item). $d_z$ = Cohen's paired effect size (small $\approx 0.2$, medium $\approx 0.5$, large $\approx 0.8$; **bolded** when $|d_z| \geq 0.4$, i.e. at least small-to-medium). $p_{\text{Holm}}$ = Holm-corrected $p$-value across the 4 contrasts within each dimension; results with $p_{\text{Holm}} < 0.05$ are considered significant.

**Study setup and raters.** Participants were recruited from Prolific and filtered to first-language English speakers currently residing in the United Kingdom or the United States, aged 20–100. Each participant rates a subset of the evaluation items, with each session consisting of 20 items (18 real items and 2 attention-check items) and taking approximately 25 minutes. The $250 \times 3 = 750$ required ratings divided by 18 real items per session yield a minimum of $\lceil \frac{750}{18} \rceil = 42$ annotator-sessions; we over-recruit to absorb attention-check failures, yielding 55 included participants. Items are assigned so that every system output receives ratings from 3 independent annotators. After excluding responses that fail attention checks or quality-control criteria, we aggregate ratings at the item level. For each source sentence and method, we average the annotator responses, so each source sentence contributes one paired data point per AuthorMix vs. baseline contrast.

**Power analysis.** To estimate the number of required measurements, we conduct a power analysis prior to conducting the study. We conduct the power analysis using G*Power 3.1 assuming two-tailed paired-samples t-tests and test against $\alpha = 0.05$. With power = .80 and a medium paired effect size of dz = .50, G*Power required 34 paired items. Our design includes 50 paired source sentences, satisfying this requirement. In the final analysis, we correct p-values across the four planned AuthorMix vs. baseline comparisons within each evaluation dimension using Holm's method. We report paired effect sizes (Cohen's $d_z$) and Holm-corrected p-values for each contrast.

**Paired t-test results.** Table 16 reports the per-dimension paired **t**-tests across the four AuthorMix vs. baseline contrasts (N = 50 paired source sentences, 3 raters per item; Holm-corrected within dimension, $\alpha = 0.05$).

**Inter-annotator agreement.** Table 17 reports Krippendorff's $\alpha$ and the mean pairwise raw agreement for each dimension, pooled across methods. We use ordinal $\alpha$ for MP and nominal $\alpha$ for Fluency and Style. Agreement is **fair** for MP ($\alpha = 0.32$), reflecting that annotators broadly converge on meaning preservation. For **Fluency**, raw agreement is high (0.70) but $\alpha$ is low (0.19); this gap is the expected paradox under a heavily skewed binary label distribution; most rewrites are judged grammatical, inflating chance agreement and depressing $\alpha$. For **Style**, $\alpha = 0.11$ is consistent with the known difficulty of fine-grained authorship discrimination for untrained raters (Patel et al., 2024; Horvitz et al., 2024) raw 2AFC agreement (0.56) sits only modestly above chance (0.50). Among the five methods, AuthorMix attains the highest per-method Style $\alpha$ (0.18) and the highest raw pairwise agreement on both MP (0.61) and Style (0.61); STYLL has the highest per-method Fluency $\alpha$ (0.22) and raw Fluency agreement (0.83).

| **Method** | **MP** (ord.) | | **Fluency** (nom.) | | **Style** (nom.) | |
|---|---|---|---|---|---|---|
| | $\alpha$ | raw | $\alpha$ | raw | $\alpha$ | raw |
| **AuthorMix** | 0.25 | **0.61** | 0.08 | **0.81** | **0.18** | **0.61** |
| ASTRAPOP-DPO | 0.21 | 0.43 | 0.05 | 0.68 | 0.14 | 0.57 |
| Few-shot | −0.03 | 0.45 | 0.00 | 0.64 | 0.06 | 0.55 |
| TinyStyler | 0.37 | 0.51 | 0.13 | 0.56 | −0.08 | 0.48 |
| STYLL | **0.38** | 0.53 | **0.22** | 0.83 | 0.12 | 0.59 |
| Pooled | 0.32 | 0.51 | 0.19 | 0.70 | 0.11 | 0.56 |

Table 17: Inter-annotator agreement on item-level ratings (N = 50 items per method, 3 raters per item; 250 items and 55 participants pooled). **raw** = mean pairwise raw agreement per item. **Bold** = best per column.

| **Method** | *Primary scoring functions (STAR + MIS)* | | | | | *Held-out scoring functions (CAV + SBERT)* | | |
|---|---|---|---|---|---|---|---|---|
| | **Toward↑** | **MIS↑** | **Joint↑** | **CoLA↑** | **Away↑** | **Toward$_{\text{CAV}}$↑** | **SBERT↑** | **Joint$_{\text{CAV}}$↑** |
| ASTRAPOP-SFT | 0.16 | 0.63 | 0.29 | 0.80 | 0.63 | 0.26 | 0.73 | 0.37 |
| ASTRAPOP-DPO | 0.17 | 0.63 | 0.29 | **0.83** | **0.78** | **0.36** | 0.71 | 0.45 |
| ASTRAPOP-GRPO | 0.17 | 0.62 | 0.30 | 0.76 | 0.62 | 0.28 | 0.72 | 0.39 |
| ASTRAPOP-GRPO (matched) | **0.18** | 0.55 | 0.29 | 0.75 | 0.63 | 0.26 | 0.68 | 0.37 |
| AuthorMix GRPO-LW ($k = 4$) | 0.16 | **0.83** | **0.34** | **0.83** | 0.67 | 0.35 | **0.84** | **0.50** |

Table 18: **ASTRAPOP-GRPO ablation.** All rows use LLaMA-3.1-8B-Instruct as the base model. ASTRAPOP-SFT and ASTRAPOP-DPO numbers carry over from Table 13 for reference. ASTRAPOP-GRPO results are evaluated $n = 100$ source–target pairs. The AuthorMix GRPO-LW row reports values at $k = 4$, selected by the primary Joint score as in Section 5. **Bold** = column max; underline = second-best.

**Human fluency vs. automatic CoLA.** TinyStyler's human fluency score (0.49) appears inconsistent with its automatic CoLA score of 0.68 in Table 13. The gap is a sampling artefact rather than a disagreement between the measures: the CoLA value in Table 13 is computed over all 1,600 test instances, whereas the human study rates a 50-sentence subset. Recomputing CoLA on exactly the 250 rated items gives 0.54 for TinyStyler, in line with the human judgement. Across all 250 items human fluency and CoLA are positively and significantly correlated (Spearman $\rho = 0.41$, $p < 0.0001$), so CoLA tracks human fluency judgements at the item level.

### C.7 ASTRAPOP-GRPO: Disentangling the Optimizer from the Adapter Mixing

This ablation asks whether AuthorMix's gain comes from the parameterization or from the optimizer. AuthorMix differs from ASTRAPOP along two axes: (i) the optimizer (GRPO vs. DPO) and (ii) the parameterization (per-target layer-wise mixing of an adapter library vs. a single global adapter trained jointly across all high-resource authors). To disentangle the contribution of these two factors, we re-train ASTRAPOP using the same SFT → GRPO pipeline as AuthorMix (Section 3.2.2), holding every other training detail constant. In this ASTRAPOP-GRPO retraining, the base model is frozen and the LoRA adapter weights are updated. In contrast, AuthorMix freezes both the base model and the entire LoRA library, training only the per-target layer-wise (or adapter-wise) mixing weights.

**Setup.** The starting checkpoint, training/validation splits, prompt template, cutoff length (256), learning rate ($5 \times 10^{-6}$), and base model (LLaMA-3.1-8B-Instruct) are kept identical to those of ASTRAPOP-DPO. The only intentional change is the policy-optimization algorithm: DPO's preference loss over chosen/rejected pairs is replaced with GRPO using the same joint reward $S(x_s, x_{s\to t}, \mathcal{X}_t) = \sqrt{T \times MIS}$ as AuthorMix (Section 3.2). All GRPO hyperparameters match the AuthorMix-GRPO sweep (`num_generations` = 16, no KL regularization, temperature 1.0, top-$p$ 0.95). We report two training budgets: 300 steps, matching the step count of AuthorMix-GRPO's layer-wise weight learning, and full training similar to ASTRAPOP-DPO. At inference, ASTRAPOP-GRPO reads the target style from the same in-prompt reference texts as ASTRAPOP-DPO, so the only difference between the two trained models is the optimizer.

**Results.** Table 18 reports the comparison. ASTRAPOP-GRPO is competitive with ASTRAPOP-DPO on the primary Joint score (0.30 vs. 0.29 for DPO), but it does not improve over DPO under the held-out CAV+SBERT scoring functions ($\text{Joint}_{\text{CAV}}$ = 0.39 vs. 0.45). At the shorter 300-step budget, the model under-trains on meaning preservation (MIS = 0.55); the longer budget recovers it only partially (MIS = 0.62), still well below AuthorMix's MIS of 0.82–0.83 across all $k$ (see Table 13).

Crucially, both ASTRAPOP-GRPO variants remain substantially below the strongest AuthorMix configuration (GRPO layer-wise at $k = 4$, selected as in Section 5 by the primary Joint score) on every metric of interest: Joint 0.30 vs. 0.34, MIS 0.62 vs. 0.83, and $\text{Joint}_{\text{CAV}}$ 0.39 vs. 0.50. The gap persists after roughly 10 × more training steps and under the held-out CAV+SBERT scoring functions. We conclude that the optimizer alone does not account for AuthorMix's advantage: granting ASTRAPOP the same optimizer, same reward, and a longer training budget still leaves a substantial margin, and this margin reflects the value of per-target layer-wise adapter mixing over a single global model.

## C.8 Signed vs. Convex Mixing Weights

CAT (Prabhakar et al., 2025) also learns one merging coefficient per layer, but constrains it to $[0, 1]$ and composes exactly two adapters (Section 3.2). Vanilla CAT cannot be run in our setting, since it learns its weights by supervised gradient descent on labelled target-task data and no gold rewrites exist for a low-resource target author. We therefore isolate the sign constraint alone: we re-run the same AuthorMix GRPO setup (Appendix B.4.3), differing **only** in projecting each layer's weight vector onto the simplex.

| Weights | $k$ | Toward↑ | MIS↑ | Joint↑ | collapsed |
|---|---|---|---|---|---|
| Convex | 2 | 0.01 | 0.27 | 0.02 | 91 / 100 |
| Convex | 3 | 0.02 | 0.28 | 0.05 | 78 / 100 |
| Convex | 4 | 0.00 | 0.23 | 0.01 | 94 / 100 |
| **Signed** | 4 | **0.16** | **0.83** | **0.34** | **0 / 100** |

Table 19: **Signed vs. convex (softmax-constrained) layer-wise mixing weights.** Signed is AuthorMix. Scores are averaged over the 100 source–target pairs × 10 evaluation runs; **collapsed** counts pairs whose Joint score falls below 0.05.

The convex variant does not converge (Table 19). At $k = 4$ it reaches a mean Joint of 0.01 against 0.34 for the signed model, with 94 of the 100 pairs and all 10 target authors below a Joint of 0.05; MIS drops to 0.23 while Away rises to ≈ 1.0, i.e. the model destroys the source text rather than restyling it. The pattern is the same at $k = 2$ and $k = 3$. The learned weights explain why: the signed solutions place consistently **negative** weights on the deep layers (24–31) (Figure 3), which a convex parameterization cannot represent, forcing positive mass onto all $k$ adapters at every layer.

## C.9 The importance of Toward and MIS

The AuthorMix reward is the geometric mean $R = T^\alpha \times M^\beta$ of the Toward style component $T$ and the MIS meaning component $M$ (Section 3.2.2), with $\alpha = \beta = 0.5$. To verify that both components are necessary, we re-trained GRPO-layerwise at $k = 4$ for Toward-only $(\alpha = 1.0, \beta = 0.0)$ and MIS-only $(\alpha = 0.0, \beta = 1.0)$ while keeping all other hyperparameters fixed. Due to the cost of running a full sweep, we evaluated each variant on a two-author target subset (Doyle, Wharton; $n = 20$ source–target pairs) and report the balanced AuthorMix configuration on the same 2 authors for a fair comparison. We report both the STAR-based primary metrics (used as the training reward) and the held-out CAV+SBERT scoring functions, to verify that the conclusion is not an artefact of the models used at training time.

Table 20 reports the outcome. The result is consistent across both evaluation pipelines. MIS-only converges to a near-identity policy: Toward drops to 0.04 under both STAR and the held-out CAV (close to the Neutral-Text reference bound of 0.01 in Table 13), as the reward has no incentive to actually shift style; the trivially high SBERT of 0.98 reflects this near-copy behaviour rather than a meaningful transfer. Toward-only optimises

| Reward $(\alpha, \beta)$ | Setting | *Primary scoring functions (STAR + MIS)* | | | | *Held-out scoring functions (CAV + SBERT)* | | |
|---|---|---|---|---|---|---|---|---|
| | | **Toward ↑** | **MIS ↑** | **Joint ↑** | **Away ↑** | **Toward$_{\text{CAV}}$ ↑** | **SBERT ↑** | **Joint$_{\text{CAV}}$ ↑** |
| $(1.0, 0.0)$ | Toward-only | 0.11 | 0.80 | 0.27 | **0.75** | **0.31** | 0.79 | 0.45 |
| $(0.0, 1.0)$ | MIS-only | 0.04 | 0.82 | 0.14 | 0.44 | 0.04 | **0.98** | 0.11 |
| $(0.5, 0.5)$ | AuthorMix (Joint) | **0.15** | **0.85** | **0.33** | 0.65 | **0.31** | 0.88 | **0.46** |

Table 20: Reward decomposition at $k = 4$ (GRPO layer-wise), evaluated on 2 target authors (Doyle, Wharton; $n = 20$ source–target pairs). **Bold** = best per column.

Toward exclusively, yet achieves a lower Toward score (0.11) than the balanced reward (0.15) under STAR, and also loses on meaning preservation (MIS = 0.80 vs. 0.85 under STAR; SBERT = 0.79 vs. 0.88 under CAV+SBERT) without MIS to anchor meaning, the policy explores degenerate rewrites that hurt Toward too. The balanced Joint reward is the only configuration that hits high marks on both axes, with the geometric-mean form penalising any drop on either component proportionally; it attains the highest Joint score under STAR (0.33 vs. 0.27 for Toward-only and 0.14 for MIS-only) and matches Toward-only on the held-out Joint$_{\text{CAV}}$ score (both 0.46) while doing so without sacrificing meaning.

### C.10 Analyzing Resource Consumption

We further provide an analysis on the resource consumption for each methods, highlighting the efficiency of AuthorMix (see Table 3). For instance, on comparable compute infrastructure, ASTRAPOP requires ∼48h of LoRA SFT + DPO training on all 10 high-resource authors (Liu et al., 2024a). TinyStyler is pre-trained using ∼8M texts from over 1M users (Horvitz et al., 2024). In contrast, AuthorMix selects a small, targeted subset of authors and adapts to any new target by learning a total of only ∼128 scalar mixing weights (k=4) which takes ∼40 minutes on a single GPU and requires no model retraining. Moreover, new adapters can be added independently to $\mathcal{A}$ with LoRA SFT requiring ∼2h of training on a single GPU.

AuthorMix provides an easily extensible alternative to large, monolithic models where producing a target style specific model in less than an hour. Our results indicate that this per-target investment leads to a genuine per-target optimization, also maximizing style transfer and meaning preservation for each target style.

**Wall-Clock Training Cost:** To facilitate fair efficiency comparisons, we report wall-clock training cost *per target author* the total compute required before the system can serve a single new target individual. This is the practically relevant unit: it measures how quickly a user can be onboarded.

For AuthorMix, the cost to adapt to one target author has two components:

1. **Adapter training (one-time).** Each *high-resource author*'s LoRA adapter is fine-tuned independently in ∼2 h on a single GPU. With $k$ related authors, this costs $k \times 2$ h. These adapters are trained once and reused for any future target that selects them, so this cost is shared across targets. For the *first* target using a given adapter set it must be paid in full; for subsequent targets that reuse the same adapters, it is zero.
2. **Weight learning (per target).** Learning the mixing weights via GRPO takes less than an hour per target author on a single GPU. This is the only cost that recurs for each new target.
3. Inference cost remains unchanged across all AuthorMix variants, regardless of $k$, since the mixed adapters are merged into the base model; it is therefore comparable to that of baselines built on the same LLaMA-3.1-8B-Instruct backbone.

The total cost for the first target at $k = 4$ is therefore $4 \times 2 + 0.67 \approx 8.7$ h. Each additional target that reuses the same adapter library requires only ∼40 min.

For the baselines, the entire training cost must be invested before *any* target can be served:

- **ASTRAPOP-SFT:** ∼10 h of full SFT fine-tuning (one-time).
- **ASTRAPOP-DPO:** ∼48 h of SFT + DPO training (one-time).
- **TinyStyler:** reconstruction pre-training on ∼8M texts, full SFT, distillation. Despite using a small 800M-parameter backbone, TinyStyler's training cost remains high (>48

| Target Style Reference texts (Fitzgerald): "That's what you ought to do," she exclaimed triumphantly.", "I can just see you," she stormed, "letting him back you down!", "Sit down and wait," suggested the lieutenant nonchalantly.... | |
|---|---|
| **Method** | **Output (target: Fitzgerald)** |
| *Source (Widger):* | 'I cannot do it,' said Ethelberta; 'and I will not.' |
| AuthorMix | "No, I won't do it," Ethelberta declared firmly. "Nor will I, under any circumstances." |
| ASTRAPOP-DPO | "I couldn't do it," Ethelberta said, "and I don't want to do it. |
| TinyStyler | Ethelberta said, "I cannot do it, and I will not do it." |
| TinyStyler-Sim | Ethelberta said, 'I cannot do it, and I will not.' |
| STYLL | Ethelberta declared she couldn't and wouldn't even try. |
| Few-shot (GPT-5.1) | "I can't do it," Ethelberta declared; "and I won't." |
| *Source (Widger):* | 'Well; better late than never, if so be he'll have her now.' |
| AuthorMix | "Better late than never, I suppose, and if he's going to have her, it's just as well he does it now." |
| ASTRAPOP-DPO | "Yes, it's better for him to have her now, even if it is late. |
| TinyStyler | 'Well, better late than never, if he has her now.' |
| TinyStyler-Sim | 'Well, better late than never, if he'll have her now, if he doesn't have her now…' |
| STYLL | (Roguishly) Better late than never, right? |
| Few-shot (GPT-5.1) | "Well, it's just as well he's come around at last—provided he's still willing to take her." |
| *Source (Widger):* | Darton held his breath till he said with treacherous solicitude, 'Try again—'tis coyness.' |
| AuthorMix | He held his breath, and then whispered with a treacherous mixture of concern and flirtation, 'Try again, it's just coyness.' |
| ASTRAPOP-DPO | Darton kept quiet until he spoke, as if she had been coy about it and he was urging her to be bold again. |
| TinyStyler | Darton was a little squirming in his seat until he said with treacherous solicitude, 'Try again.' |
| TinyStyler-Sim | Darton held his breath until he said with a treacherous solicitude, "Try again—'tis coyness." |
| STYLL | Darton kept mum until he finally broke the silence, hinting that their hesitance was just a coy act, and egging them on to have another go. |
| Few-shot (GPT-5.1) | Darton, still holding his breath, murmured with a deceitful gentleness, "Try once more—it's only shyness." |
| *Source (Widger):* | One of them actually said that you must be fifty to have got such an experience. |
| AuthorMix | "If you've had an experience like that, you must be at least fifty," he said bluntly. |
| ASTRAPOP-DPO | "He said a man's got to be fifty years old to have that much experience." |
| TinyStyler | One of them said you had to be fifty to have such an experience, smh. |
| TinyStyler-Sim | One of them said that you must be 50 to have had such an experience … |
| STYLL | It was said that you needed to be half a century old to have gathered so much experience. |
| Few-shot (GPT-5.1) | One of them even ventured the opinion that I must be fifty years old to have managed such an adventure. |

Table 21: Multiple source sentences rewritten into Fitzgerald's style by all methods. AuthorMix produces stylistically distinct outputs while preserving meaning across different source inputs for the same target author.

GPU h) because it requires full-model fine-tuning (reconstruction plus self-distillation) and is likely higher than both ASTRAPOP and our method.

- **STYLL / Few-shot:** No training cost (inference-only via API).

Once trained, baseline methods adapt to new targets at zero additional cost (by changing the prompt or embedding). AuthorMix instead pays $\sim$ 40 min per new target but requires **5.5× less compute** than ASTRAPOP-DPO and far less dataset than TinyStyler to reach the first target.

## C.11 Output Examples

### C.11.1 MIS output analysis

Table 1 shows outputs of different style-transfer methods for a single sentence along with their mutual implication score (MIS) that captures how well the meaning is preserved. The original conveys three key elements: the speaker ("I"), the intent ("to tell something"), and the addressee ("you all"). Only AuthorMix preserves all three. ASTRAPOP changes the speaker to "he" and drops the group address; STYLL preserves the speaker but shifts "you all" to "everything," losing the direct address; TinyStyler preserves the content but degenerates into ungrammatical repetition. GPT-5.1 retains all three elements but adds embellishment not present in the original ("I broke in at last," "lay it bare"). The neutral baseline depersonalizes the speaker entirely ("the speaker," "they") and removes the addressee.

## C.12 Qualitative Examples

Table 21 by fixing a single target author (Fitzgerald) and varying the source text, revealing how each method handles different inputs when targeting the same style.

| Failure type (pair, sent) | Source → AuthorMix output | MIS | Toward |
|---|---|---|---|
| Foreign-word translation (Brontë→Christie) | **SRC:** You remember what we saw that night in the berceau?<br>**OUT:** Do you remember what we saw that night in the **cradle**? | **0.03** | 0.08 |
| Style under-shift (Baum→Kipling) | **SRC:** Thoughts should be restrained in the same way as your oil, and only applied when necessary, and for a good purpose.<br>**OUT:** Use your thoughts as you would your oil — sparingly and only when the engine demands it, and for a good reason. | 0.78 | **0.00** |
| Content drop on a long source (Wheatley→Christie) | **SRC:** The remarks with which Dr. Johnson opens the preface to his English Dictionary may well be applied to the Indexer: "It is the fate of those who toil at the lower employments of life to be rather driven by the fear of evil…"<br>**OUT:** The Indexer is driven by the fear of errors, rather than attracted by the prospect of praise, and is exposed to censure without hope of recognition. | **0.12** | **0.07** |
| Intermittent empty output (Baum→Christie) | **SRC:** Ozma was studying a roll of manuscript which she had taken from the Royal Library…<br>**OUT:** (empty) | **0.00** | **0.00** |
| Near-identity copy on verse (Carroll→Doyle) | **SRC:** "And will it do, O will it do To take them in a lump- As 'the wild man went his weary way To a strange and lonely pump'?" "Nay, nay!"<br>**OUT:** "Will it do, O will it do, to take them in a lump, As the wild man went his weary way to a strange and lonely pump? Nay, nay!" | 0.95 | **0.00** |

Table 22: Five representative AuthorMix failure modes. MIS is the raw NLI score for the displayed pair; Toward is the scaled STAR-based score (so 0 means no movement past the source). **Bold** marks the score(s) that diagnose the failure for each row: (i) and (iii) are caught by MIS; (ii) and (v) are invisible on MIS alone and are caught only by Toward; (iv) registers on both axes simultaneously.

### C.12.1 Failure Modes and Bad Examples

We surface five representative AuthorMix failure cases in Table 22, each from a different (source, target) pair under the GRPO layerwise $k = 4$ configuration. Together they illustrate why both axes need to be reported: each case is invisible on at least one of MIS or Toward alone.

**Discussion.** Each case requires a different signal to diagnose: **(i) Brontë→Christie.** The model translates the French loan-word `berceau` into its English gloss `cradle`. The NLI-based MIS rejects this synonym substitution as a meaning change, collapsing to 0.03; the Toward score is paradoxically slightly higher (0.08) than for baselines that copy the loan word, since the resulting sentence reads more uniformly English. MIS is the diagnostic here. **(ii) Baum→Kipling.** The rewrite is a fluent paraphrase (MIS = 0.78) but does not shift toward Kipling's style (Toward = 0.00). MIS alone would suggest the rewrite is fine; Toward captures the under-shift. **(iii) Wheatley→Christie.** The rewrite strips most of the embedded Dr. Johnson quote, collapsing MIS to 0.12. MIS is the unmistakable diagnostic here; the low Toward (0.07) corroborates that the rewrite has also not landed in Christie's style. **(iv) Baum→Christie.** An empty output where both axes drop to zero simultaneously. The case is rare (a small fraction of seeds for that specific input) and is already absorbed by the ten-seed averaging protocol. **(v) Carroll→Doyle.** The verse is reproduced almost verbatim (MIS = 0.95) with no movement toward Doyle (Toward = 0.00). MIS alone would miss this failure entirely; only Toward exposes it. Cases (ii) and (v) are MIS-misses; cases (i) and (iii) are MIS-catches; case (iv) registers on both. Reporting MIS and Toward jointly is therefore essential: neither axis alone catches all five failure modes.

### C.13 Human Evaluation Materials

Table 23 reproduces the three questions shown to raters for every item, and Table 24 summarises the overall study plan. Interface screenshots are shown in Figure 14, Figure 15, Figure 16, and Figure 17.

| Question | Prompt and options shown to the rater |
|---|---|
| Q1: Meaning | *How well does the rewritten text preserve the meaning of the original?*<br>• Fully preserved — same content.<br>• Partially preserved — some content changed, lost, or added.<br>• Substantially changed — meaning is largely different or lost.<br>(ordinal, scored 2 / 1 / 0) |
| Q2: Fluency | *Is the rewritten text fluent and grammatical?*<br>• Yes — reads naturally, no grammar issues.<br>• No — has grammar errors or reads awkwardly.<br>(binary, scored 1 / 0) |
| Q3: Style match | *Which author's style does the rewritten text more closely match?* Rater is shown 5 example sentences from each of two authors (labelled "Author A" and "Author B") and chooses one. Instruction emphasises focusing on *how* things are written (word choice, sentence structure, punctuation), not *what* is written. Optional free-text justification. (2AFC) |

Table 23: Per-item instructions and response options shown to Prolific raters. Author A / Author B assignment to source / target is randomised per item to control for position bias.

| Field | Value |
|---|---|
| Study title | Evaluating Text Rewrites: Meaning, Fluency, and Writing Style |
| Platform | Prolific |
| Rater pool | First-language English speakers in UK/US, aged 20–100 |
| Session length | 20 items per session (18 real + 2 attention checks), ∼25 min, one sitting |
| Reward | £4.79 per session (£11.49/hr) + bonus for optional Q3 comments |
| Items per system output | 3 independent raters |
| Source corpus | Project Gutenberg 19th-/early-20th-century English novels |
| Devices allowed | Desktop, mobile, tablet |
| Content warning shown | Archaic language, period-specific cultural/religious framings, references to violence in narrative context; nothing explicit or graphic |
| Workflow | Enter Prolific ID → instructions + 2 worked examples → 20 items (no back-navigation) → completion code |

Table 24: Human-evaluation study plan (Prolific-facing setup).

In this short study, you will be shown 20 short items, one at a time. Each item contains:

1 . An **original sentence** and a **rewritten version** of that sentence.
2 . A small set of **example sentences from two different authors** (labelled "Author A" and "Author B").

For each item, you will answer **3 questions**:

- **Meaning preservation** — How well does the rewrite preserve the meaning of the original? *(Fully preserved / Partially preserved / Substantially changed)*
- **Fluency** — Is the rewritten text grammatical and natural-sounding English? *(Yes / No)*
- **Style match** — Looking at the two sets of author examples, which author's writing style does the rewrite more closely match? *(Author A / Author B)*

This study uses excerpts from **classic English literature (19th- and early-20th-century novels from Project Gutenberg)**. Some passages may contain:

- **Archaic language** that can read differently to modern audiences (e.g., older usages of words like *savage* or *queer*).
- **Period-specific cultural and religious framings** typical of the era's literature.
- **References to violence in narrative context**, such as descriptions of battle wounds.

None of the content is explicit, graphic, or intended to shock. **If you prefer not to engage with historical literary material of this kind, please do not take part.**

Before starting, you will see brief instructions with **two worked examples**.

You will enter your **Prolific ID** at the start, and at the end you will receive a **completion code** to paste back into Prolific to confirm your submission. **You cannot go back to previous items, so please consider each answer before clicking "Next".**

**Estimated time:** ~25 minutes. Please complete in one sitting.

> This study includes sensitive topics. Learn more about study content warnings here.
>
> **Additional details:**
> Excerpts from 19th-century English novels. Contains archaic language, period-specific cultural/religious framings, and references to violence (e.g., battle wounds). Nothing explicit or graphic.

Devices you can use to take this study:

Desktop Mobile Tablet

Figure 14: Study Description

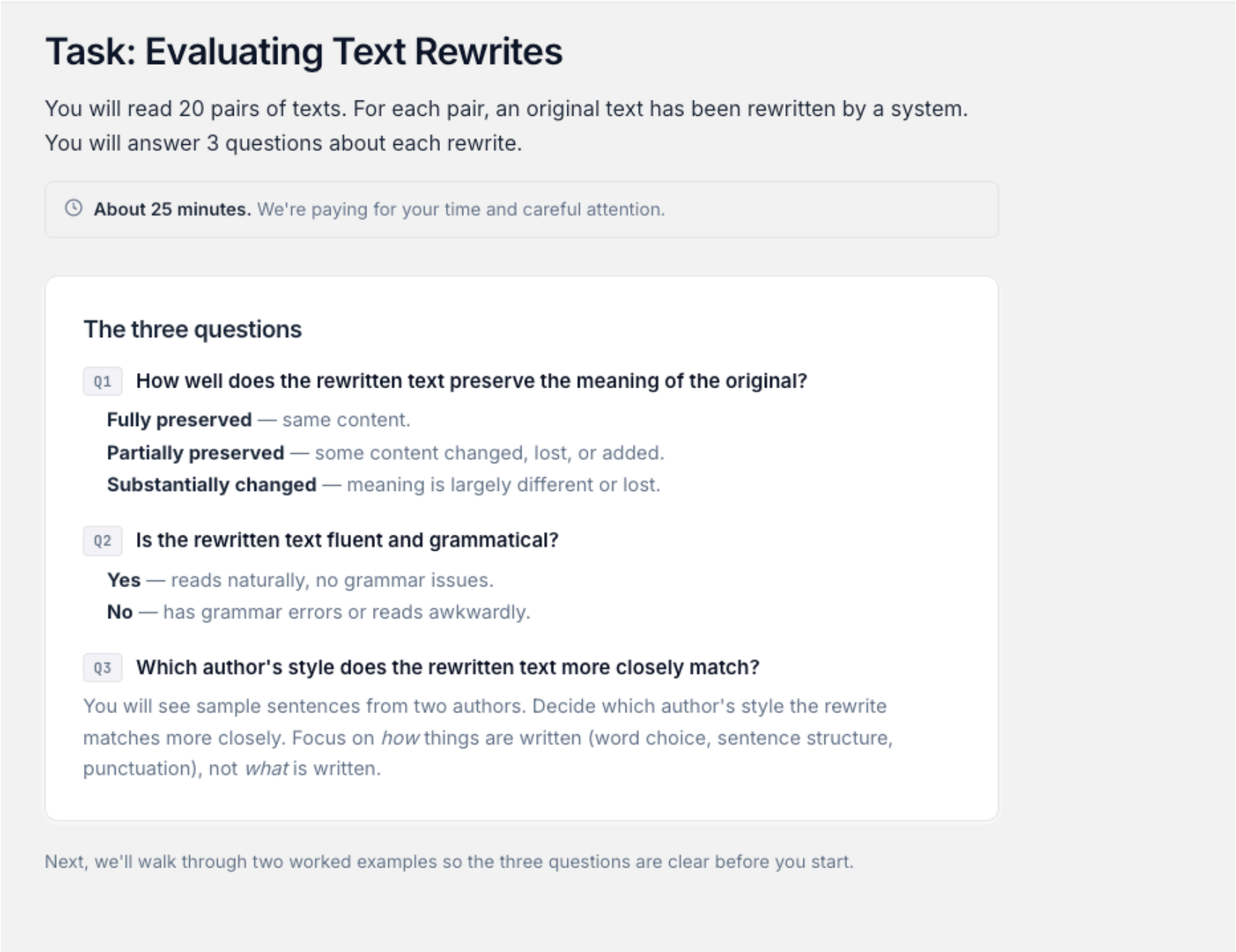

Figure 15: Task Overview

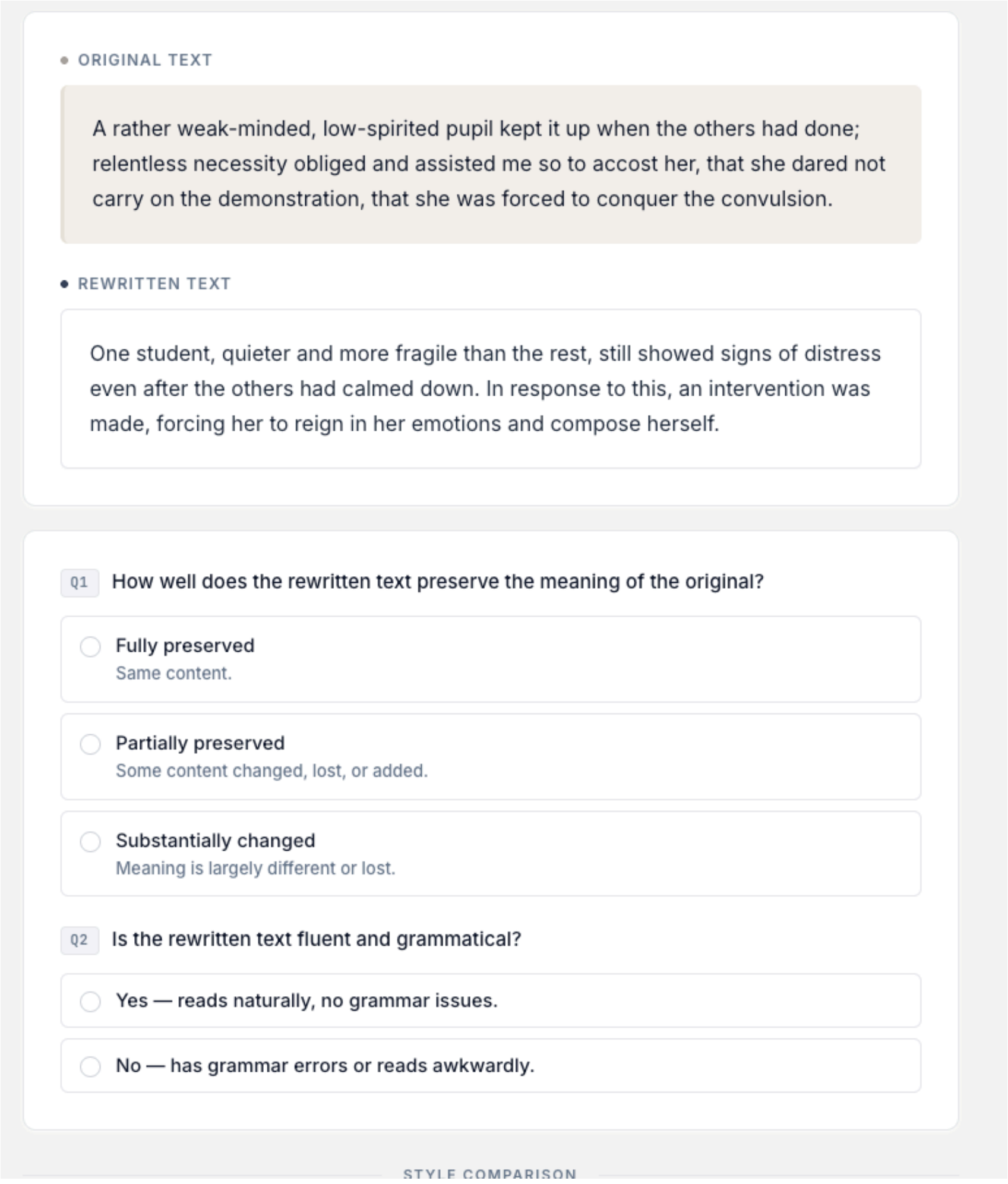

Figure 16: Per-item interface for **Q1 (Meaning preservation)** and **Q2 (Fluency)**: the original sentence and its rewrite are shown side by side, and the rater answers an ordinal 3-way meaning question and a binary fluency question.

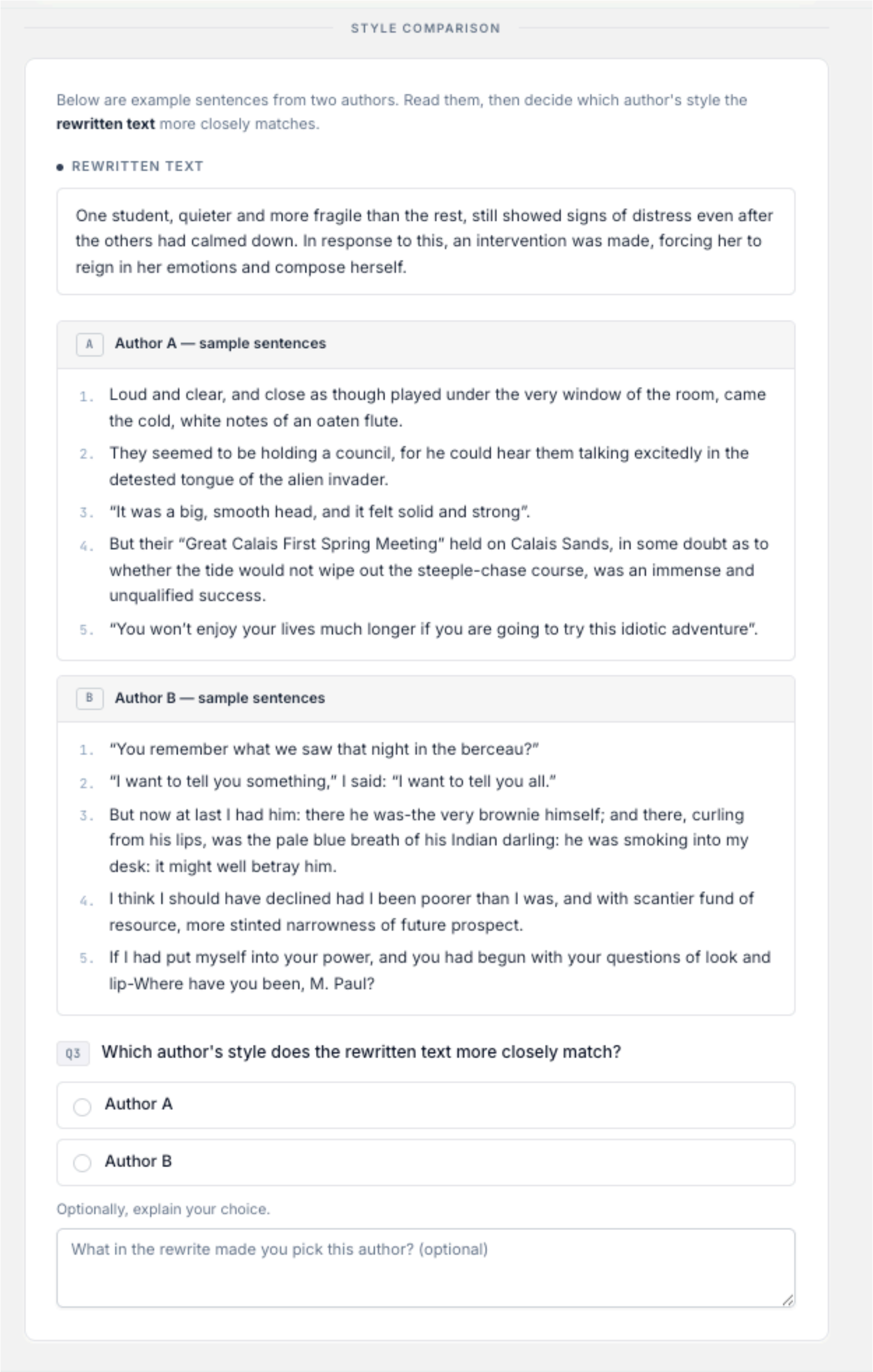

Figure 17: Per-item interface for **Q3 (Style match)**: the rewrite is shown alongside five reference sentences from each of two authors (labelled "Author A" and "Author B", randomised per item), and the rater selects which author's style the rewrite more closely matches. An optional free-text box invites a brief justification.